\documentclass[letterpaper, 10pt, conference]{ieeeconf}  

\IEEEoverridecommandlockouts                              
\usepackage{lipsum} 
\usepackage{scrextend} 
\usepackage[pdftex]{graphicx} 
\usepackage[caption=false]{subfig} 
\usepackage{siunitx} 
\usepackage{multirow} 
\usepackage{array} 
\let\labelindent\relax\usepackage{enumitem} 
\usepackage{color} 
\usepackage{colortbl} 
\usepackage{pifont} 
\usepackage{amsmath} 
\usepackage{amssymb}
\usepackage{dblfloatfix} 
\usepackage{tikz}
\usepackage{tikz-uml} 
\usetikzlibrary{shapes, arrows, fit, calc, positioning, automata, decorations.markings, backgrounds}
\tikzset{>=latex}

\makeatletter
\let\NAT@parse\undefined
\makeatother
\PassOptionsToPackage{hyphens}{url}
\usepackage{hyperref} 
\hypersetup{colorlinks=true,linkcolor=blue,urlcolor=blue,citecolor=blue,anchorcolor=blue}

\graphicspath{{images/}} 
\DeclareGraphicsExtensions{.pdf,.jpg,.png} 

\newtheorem{remark}{Remark}

\newcommand{\cmark}{\textcolor{green}{\ding{52}}} 
\newcommand{\xmark}{\textcolor{red}{\ding{56}}} 

\newcounter{spcounter}\renewcommand{\thespcounter}{\textit{\alph{spcounter}}}
\newcommand{\speciallabel}[1]{\refstepcounter{spcounter}\textsuperscript{\thespcounter}\label{#1}}
\newcommand{\specialref}[1]{\textsuperscript{\ref{#1}}}

\title{\LARGE \bf
{\tt anafi\_ros}: from Off-the-Shelf Drones to Research Platforms
}

\author{Andriy Sarabakha$^1$
\thanks{This research was supported by the NTU Presidential Postdoctoral Fellowship (award number 021820-00001).}
\thanks{$^1$Andriy Sarabakha is with the School of Electrical and Electronic Engineering~(EEE), Nanyang Technological University~(NTU), Singapore, 639798. {\tt\small andriy.sarabakha@ntu.edu.sg}}%
}

\begin{document}

\maketitle

\thispagestyle{plain}
\pagestyle{plain}

\bstctlcite{IEEEexample:BSTcontrol}


\renewcommand{\thefootnote}{\fnsymbol{footnote}}

\begin{abstract}
The off-the-shelf drones are simple to operate and easy to maintain aerial systems. However, due to proprietary flight software, these drones usually do not provide any open-source interface which can enable them for autonomous flight in research or teaching. This work introduces a package for ROS1 and ROS2 for straightforward interfacing with off-the-shelf drones from the Parrot ANAFI family.  The developed ROS package is hardware agnostic, allowing connecting seamlessly to all four supported drone models. This framework can connect with the same ease to a single drone or a team of drones from the same ground station. The developed package was intensively tested at the limits of the drones' capabilities and thoughtfully documented to facilitate its use by other research groups worldwide.
\end{abstract}

\section{Introduction}


As one of the fastest-growing fields in the aerospace industry, unmanned aerial vehicles (UAVs) can provide a cost-efficient solution to many time-consuming tasks, such as subterranean exploration~\cite{Tranzatto2022SR}, 
power-line inspection~\cite{Suarez2022ICUAS} and additive manufacturing~\cite{Zhang2022Nature}. Different applications require drones equipped with appropriate sensors, for example, a thermal camera for wildfire monitoring~\cite{Valero2021EO}, a camera with a high optical zoom for aerial observation~\cite{Chen2022EO}, or a stereo camera for visual odometry~\cite{Pham2022RAL}. Custom-built drones can be designed and manufactured to meet specific needs and requirements, allowing for more flexibility in their functionality~\cite{Foehn2022SR}. However, assembling and configuring custom-made drones require technical expertise and labour time.


Unlike custom-made drones, off-the-shelf drones are readily available and can be easily purchased from many retailers. This makes them an attractive option for individuals and organisations that want to use drones for a variety of purposes, such as teaching and research. Another advantage of using off-the-shelf drones is their reliability and durability since they are built to operate in a wide range of conditions.

\newcommand{\TwoFiguresWidth}{0.5}
\begin{figure}[!t]
\centering
\subfloat[ANAFI 4K.]{\includegraphics[width=\TwoFiguresWidth\columnwidth]{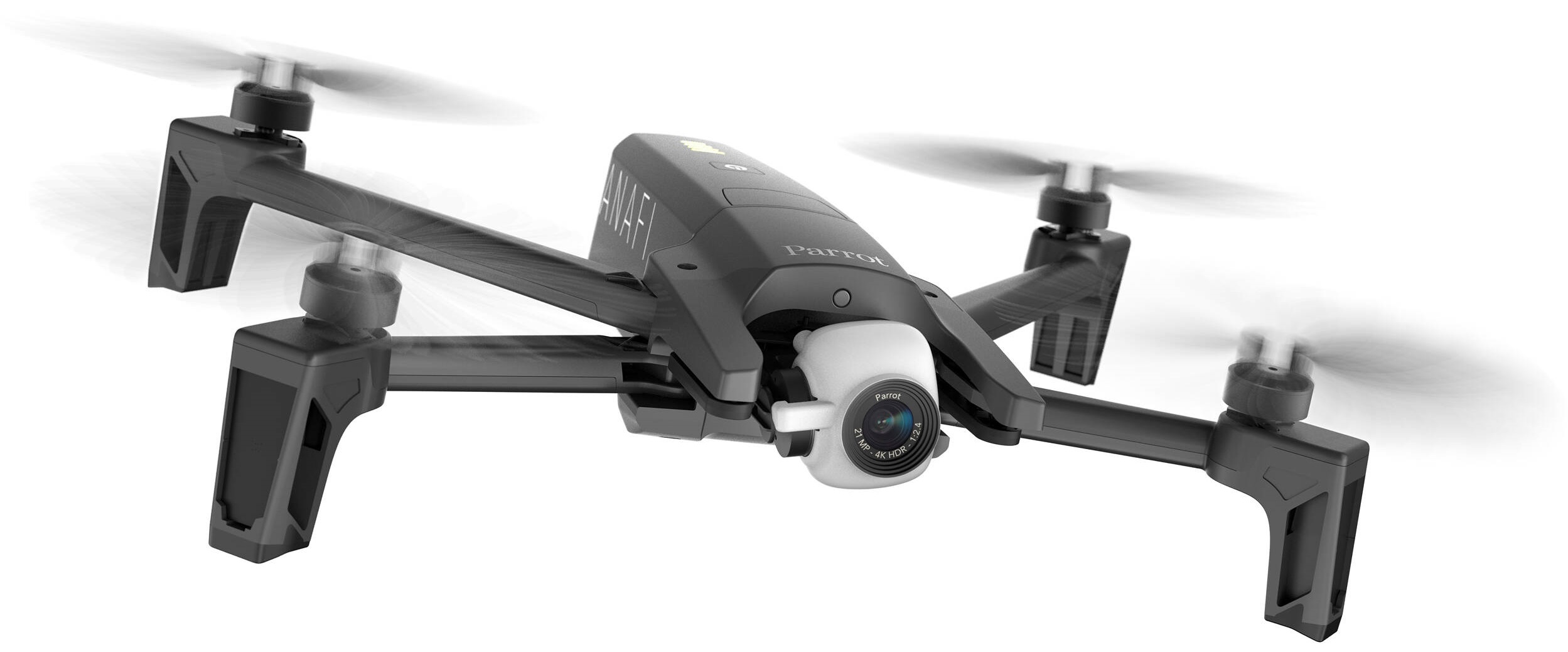}%
\label{fig:anafi_4k}} \hfill
\subfloat[ANAFI Thermal.]{\includegraphics[width=\TwoFiguresWidth\columnwidth]{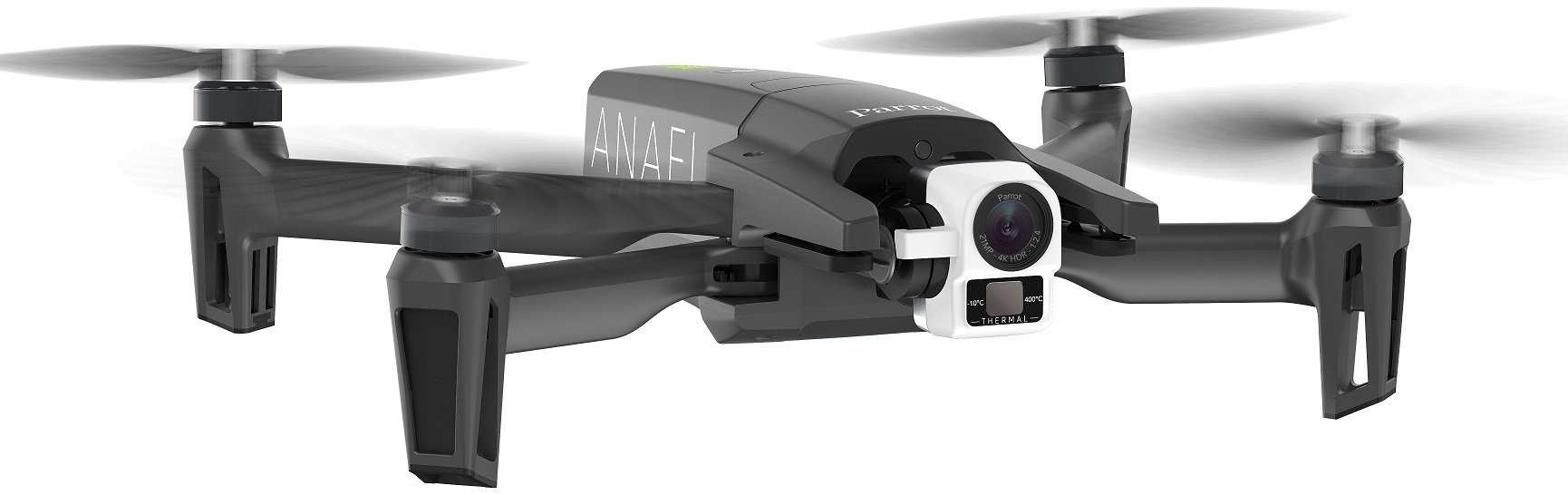}%
\label{fig:anafi_thermal}} \\
\subfloat[ANAFI USA.]{\includegraphics[width=\TwoFiguresWidth\columnwidth]{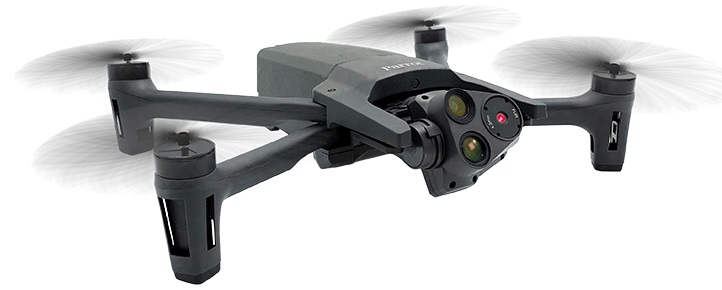}%
\label{fig:anafi_usa}} \hfill
\subfloat[ANAFI Ai.]{\includegraphics[width=\TwoFiguresWidth\columnwidth]{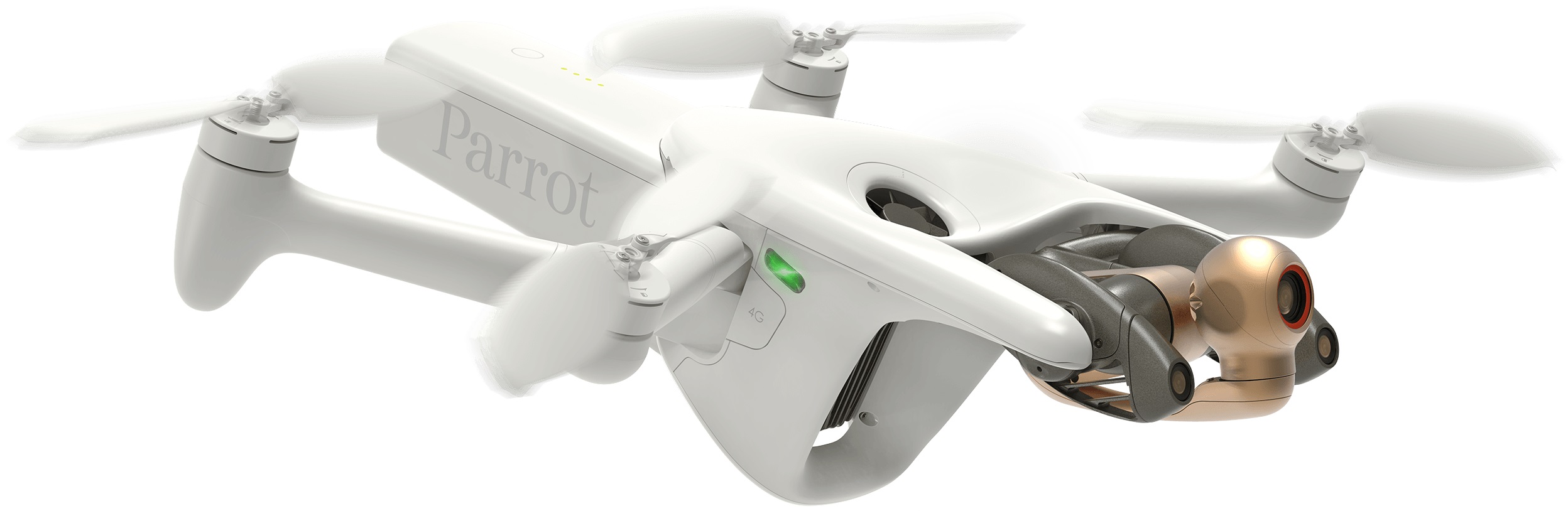}%
\label{fig:anafi_ai}}
\caption{Parrot ANAFI family drones.}
\label{fig:anafi}
\end{figure}


While there are many advantages of using off-the-shelf drones, their main limitation is the possibility of autonomous deployments, making them less suitable for certain applications. To partially overcome this issue, proprietary software development kits~(SDKs) were released by some drone manufacturers, such as DJI~\cite{dji}, Ryze~\cite{ryze}, Parrot~\cite{parrot} and Bitcraze~\cite{bitcraze}. Still, only discontinued Parrot Bebop~\cite{bebop-autonomy}, and tiny-size Ryze Tello~\cite{tello-ros} and Bitcraze Crazyflie~\cite{Preiss2017ICRA} have the interface to bridge them with the robot operating system~(ROS).
Nowadays, ROS has become a standard development environment for modern roboticists.
The main advantage of ROS is the possibility for the modularization of software, making it easy to reuse and modify individual components. Besides, ROS provides a standard set of libraries, tools and conventions for communication between different parts of a robotic system. Moreover, ROS has a large and active community with many available resources.
A huge variety of ROS packages is available for building the navigation stack for aerial robots, like visual-inertial localisation~\cite{Mur2015TRo}, environment perception~\cite{Tallamraju2019RAL}, motion planning~\cite{Zhou2021TRo}, model-based~\cite{Torrente2021RAL} and model-free~\cite{Sarabakha2020TFS} control.


This work introduces a bridge which allows a straightforward connection between the drones of the Parrot\footnote{The author has no relations or conflicts of interest with Parrot SA.} ANAFI family (illustrated in Fig.~\ref{fig:anafi}) and both ROS1\footnote{The open-source code is available at \url{https://github.com/andriyukr/anafi_ros/tree/ros1}.} and ROS2\footnote{\label{anafi-ros2}The open-source code is available at \url{https://github.com/andriyukr/anafi_ros/tree/ros2}.}. Parrot ANAFI drones were chosen because each model in the family offers unique features making them suitable for various applications. A comprehensive comparison of the characteristics of Parrot ANAFI drones is provided. The developed ROS package is hardware agnostic, allowing connecting seamlessly to all supported drones. This framework also allows connecting to single or multiple drones from the same ground station. The developed package was intensively tested at the limits of the drones and thoughtfully documented to facilitate its use by other researchers.


This work is organised as follows. First, the technical details of the experimental platforms are provided in Section~\ref{sec:platform}. Next, Section~\ref{sec:framework} describes the structure of the developed framework. Then, Section~\ref{sec:validation} provides experimental validation of the developed framework. Finally, Section~\ref{sec:conclusion} summarises this work with conclusions and future work.

\section{Experimental Platforms}
\label{sec:platform}

\begin{figure}[!b]
\centering
\includegraphics[width=0.99\columnwidth]{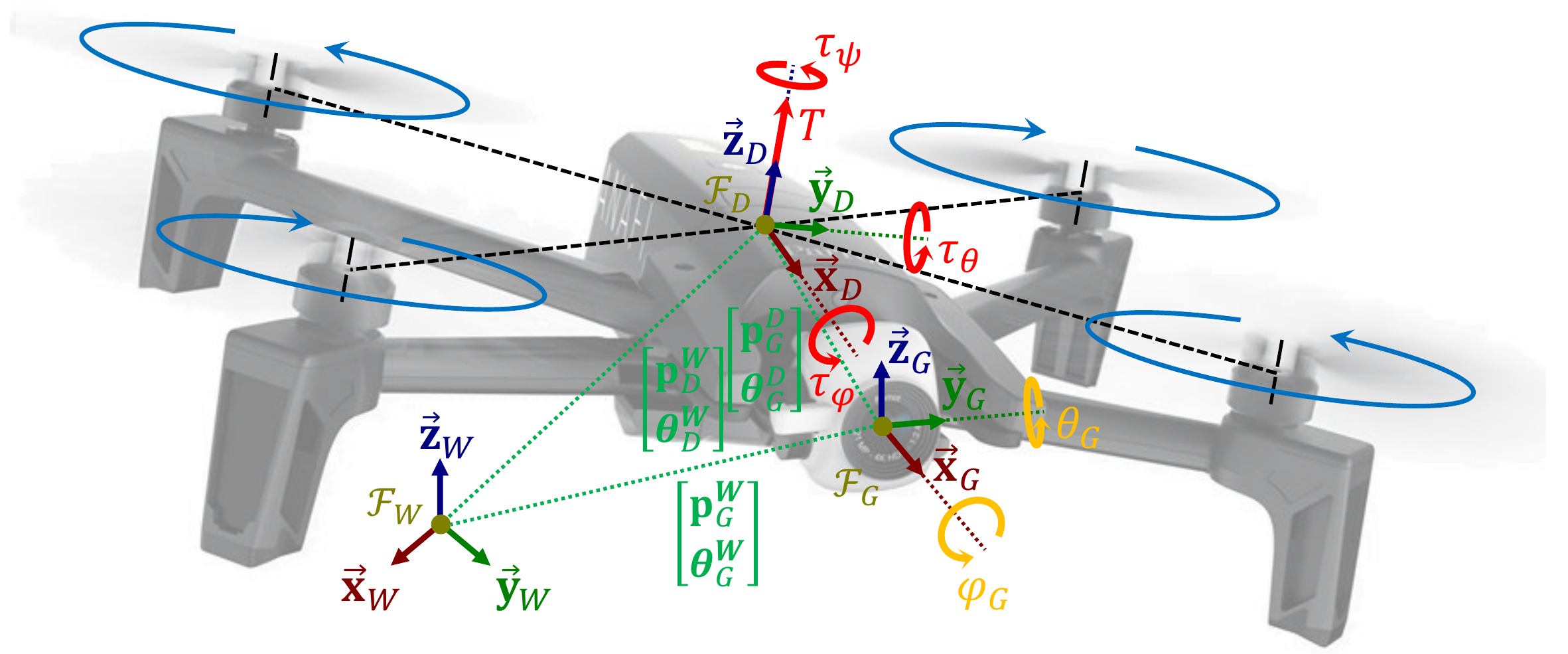}
\caption{Configuration of Parrot ANAFI with its reference frames.}
\label{fig:frames}
\end{figure}

Let the world fixed frame be $\mathcal{F}_W = \{\vec{\mathbf{x}}_W, \vec{\mathbf{y}}_W, \vec{\mathbf{z}}_W\}$, and the drone body frame be $\mathcal{F}_D = \{ \vec{\mathbf{x}}_D, \vec{\mathbf{y}}_D, \vec{\mathbf{z}}_D \}$. The origin of the body frame is located at the centre of mass (COM) of the UAV. The configuration with the corresponding reference frames is illustrated in Fig.~\ref{fig:frames}.

The absolute position of UAV $\mathbf{p}^W_D = \begin{bmatrix} x & y & z \end{bmatrix}^T$ is described by three Cartesian coordinates of its COM in $\mathcal{F}_W$. While the attitude of UAV $\boldsymbol{\theta}^W_D = \begin{bmatrix} \phi & \theta & \psi \end{bmatrix}^T$ is described by three Euler's angles: roll $\phi$, pitch $\theta$ and yaw $\psi$.

The time derivative of the position ($x$, $y$, $z$) gives the linear velocity of the UAV's COM expressed in $\mathcal{F}_W$:
\begin{equation}
\mathbf{v} = \begin{bmatrix}\dot{x} & \dot{y} & \dot{z}\end{bmatrix}^T,
\end{equation}
and the velocity expressed in $\mathcal{F}_D$ is
\begin{equation}
\mathbf{v}_D = \begin{bmatrix} v_x & v_y & v_z \end{bmatrix}^T.
\end{equation}
The relation between $\mathbf{v}$ and $\mathbf{v}_B$ is given by
\begin{equation}
\mathbf{v} = \mathbf{R}(\phi, \theta, \psi) \mathbf{v}_D,
\label{eq:linear_velocity}
\end{equation}
in which $\mathbf{R}(\phi, \theta, \psi) \in \mathsf{SO}(3)$ is the rotation matrix from $\mathcal{F}_B$ to $\mathcal{F}_W$:
\begin{equation}
\mathbf{R}(\phi, \theta, \psi) = \begin{bmatrix} \text{c}_\psi \text{c}_\theta & \text{c}_\psi \text{s}_\phi \text{s}_\theta - \text{c}_\phi \text{s}_\psi & \text{s}_\phi \text{s}_\psi + \text{c}_\phi \text{c}_\psi \text{s}_\theta \\ \text{c}_\theta \text{s}_\psi & \text{c}_\phi \text{c}_\psi + \text{s}_\phi \text{s}_\psi \text{s}_\theta & \text{c}_\phi \text{s}_\psi \text{s}_\theta - \text{c}_\psi \text{s}_\phi \\ -\text{s}_\theta\ & \text{c}_\theta \text{s}_\phi & \text{c}_\phi \text{c}_\theta \end{bmatrix},
\label{eq:rotation_matrix}
\end{equation}
in which $\text{c}_\star$ and $\text{s}_\star$ are $\cos(\star)$ and $~\sin(\star)$, respectively.

The time derivative of the attitude $(\phi, \theta, \psi)$ gives the angular velocity expressed in $\mathcal{F}_W$:
\begin{equation}
\boldsymbol{\omega} = \begin{bmatrix}\dot{\phi} & \dot{\theta} & \dot{\psi}\end{bmatrix}^T,
\end{equation}
and the angular velocity expressed in $\mathcal{F}_D$ is
\begin{equation}
\boldsymbol{\omega}_D = \begin{bmatrix} \omega_{\phi} & \omega_{\theta} & \omega_{\psi} \end{bmatrix}^{T}.
\end{equation}
The relation between $\boldsymbol{\omega}$ and $\boldsymbol{\omega}_D$ is given by
\begin{equation}
\boldsymbol{\omega} = \mathbf{T} \boldsymbol{\omega}_D,
\label{eq:angular_velocity}
\end{equation}
in which $\mathbf{T}$ is the transformation matrix:
\begin{equation}
\mathbf{T} = \begin{bmatrix} 1 & ~\sin\phi \tan\theta & \cos\phi \tan\theta \\ 0 & \cos\phi & -~\sin\phi \\ 0 & ~\sin\phi \sec\theta & \cos\phi \sec\theta \end{bmatrix}.
\end{equation}

The vector of control inputs $\mathbf{u}$ is considered as in~\cite{Sarabakha2016CDC}:
\begin{equation}
\mathbf{u} = \begin{bmatrix} T^* & \tau_{\phi}^* & \tau_{\theta}^* & \tau_{\psi}^* \end{bmatrix}^T,
\label{eq:inputs}
\end{equation}
where $T^*$ is the reference thrust acting along $\vec{\mathbf{z}}_D$ axis, whereas $\tau_{\phi}^*$, $\tau_{\theta}^*$ and $\tau_{\psi}^*$ are the reference moments acting around $\vec{\mathbf{x}}_D$, $\vec{\mathbf{y}}_D$ and $\vec{\mathbf{z}}_D$ axes, respectively.

For mobile robots equipped with a camera mounted on a gimbal, there is also a gimbal reference frame $\mathcal{F}_G = \{\vec{\mathbf{x}}_G, \vec{\mathbf{y}}_G, \vec{\mathbf{z}}_G\}$. The origin of the gimbal frame is located in front of the UAV's COM. The position of the camera in $\mathcal{F}_D$ is $\mathbf{p}^D_G = \begin{bmatrix} x_G & y_G & z_G \end{bmatrix}^T$, while the attitude of the gimbal in $\mathcal{F}_D$ is $\boldsymbol{\theta}^D_G = \begin{bmatrix} \phi_G & \theta_G & \psi_G \end{bmatrix}^T$. The rotation matrix of the gimbal in $\mathcal{F}_W$ can be calculated with
\begin{equation}
\mathbf{R}^W_G = \mathbf{R}(\phi, \theta, \psi) \mathbf{R}(\phi_G, \theta_G, \psi_G),
\label{eq:R_world_gimbal}
\end{equation}
while the position of the camera in $\mathcal{F}_W$ can be obtained with
\begin{equation}
\mathbf{p}^W_G = \mathbf{p}^W_D + \mathbf{R}(\phi, \theta, \psi) \mathbf{p}^D_G.
\label{eq:p_world_gimbal}
\end{equation}
\subsection{ANAFI Drones}

Parrot ANAFI drones are small, lightweight UAVs mainly designed for aerial photography and videography. They are equipped with high-resolution high-dynamic-range~(HDR) cameras mounted on a $2$-axis gimbal, allowing to capture smooth and stable footage from the air. These drones have a unique folding design, making them portable and easily deployable: it takes less than $1~\si{min}$ to unfold the drone, turn it on, connect to the remote controller and take off. Moreover, ANAFI drones have reasonably long flight duration thanks to the lithium polymer (LiPo) battery, which has a built-in \mbox{USB-C} port for hassle-free charging. The Parrot ANAFI family has four drone models: basic \textit{ANAFI~4K}, \textit{ANAFI~Thermal} with a thermal camera, water- and dust-resistant \textit{ANAFI~USA} and \textit{ANAFI~Ai} with onboard computing and obstacle avoidance capabilities. These drones are illustrated in Fig.~\ref{fig:anafi}, while Table~\ref{tab:anafi} summarises the properties of each model.

\begin{remark}
Since low-level stabilisation controllers as in~\cite{Mistler2001ROMAN} 
are included in ANAFI's autopilot as illustrated in Fig.~\ref{fig:controller}, the virtual control inputs in~\eqref{eq:inputs} can be considered as:
\begin{equation}
\mathbf{u} = \begin{bmatrix} v_z^* & \phi^* & \theta^* & \omega_{\psi}^* \end{bmatrix}^T.
\label{eq:u_virtual}
\end{equation}
\end{remark}

\tikzstyle{int}=[draw, align=center, fill=blue!0, minimum size=2em, text width=1cm]
\tikzstyle{sum}=[draw, fill=blue!0, shape=circle, inner sep=0.3pt]
\begin{figure}[!b]
\centering
    \resizebox{0.99\columnwidth}{!}{
    \begin{tikzpicture}[node distance=1cm, auto, >=latex', line width=0.5mm]
        \node[int, text width=2cm, minimum height=1cm, fill=red!20](vertical){Vertical velocity controller};
        \node[int, below right=0.5cm and -1.25cm of vertical, text width=2cm, minimum height=1cm, fill=green!20](attitude){Roll \& pitch controller};
        \node[int, below right=0.5cm and -1.25cm of attitude, text width=2cm, minimum height=1cm, fill=blue!20](yaw){Yaw rate controller};
        \node[int, right=2cm of attitude, text width=2cm, minimum height=4cm, fill=cyan!20](drone){Drone dynamics};
   
        \draw[->] ($(vertical.west) + (-1cm, 0)$) -- node[pos=0.3]{$v_z^*$} (vertical.west);
        \draw[->] ($(attitude.west) + (-2cm, 0.3cm)$) -- node[pos=0.2]{$\phi^*$} ++(2cm, 0);
        \draw[->] ($(attitude.west) + (-2cm, -0.3cm)$) -- node[pos=0.2]{$\theta^*$} ++(2cm, 0);
        \draw[->] ($(yaw.west) + (-3cm, 0)$) -- node[pos=0.1]{$\omega_{\psi}^*$} (yaw.west);
        \draw[->] (vertical.east) -- node{$T^*$} ++(3cm, 0);
        \draw[->] ($(attitude.east) + (0, 0.3cm)$) -- node{$\tau_{\phi}^*$} ++(2cm, 0);
        \draw[->] ($(attitude.east) + (0, -0.3cm)$) -- node{$\tau_{\theta}^*$} ++(2cm, 0);
        \draw[->] (yaw.east) -- node{$\tau_{\psi}^*$} ++(1cm, 0);
        \draw[->] ($(drone.south) + (-0.8cm, 0)$) -- node[right, pos=0.7]{$\omega_{\psi}$} ++(0, -0.5cm) -| (yaw.south);
        \draw[->] (drone.south) -- node[right, pos=0.3]{$\boldsymbol{\theta}^W_D$} ++(0, -1cm) -| ($(attitude.south) + (-0.5cm, 0)$);      
        \draw[->] ($(drone.south) + (0.8cm, 0)$) -- node[right, pos=0.2]{$v_z$} ++(0, -1.5cm) -| ($(vertical.south) + (-0.5cm, 0)$);
        \draw[->] (drone.east) -- node{$\mathbf{p}^W_D$} ++(1cm, 0);
    \end{tikzpicture}}
\caption{Control scheme of Parrot ANAFI UAV.}
\label{fig:controller}
\end{figure}
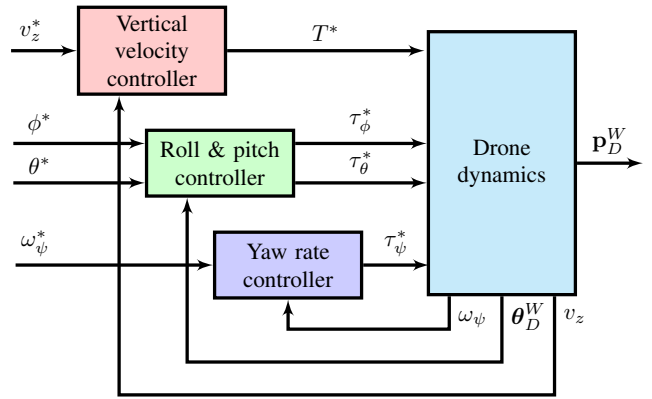

\begin{table*}[!bt]
\caption{Technical specifications of Parrot ANAFI Drones.}
\centering
\begin{tabular}{|c|l||c|c|c|c|} \hline
\multicolumn{2}{|c||}{\multirow{2}{*}{\textbf{Parameter}}}                                                                  & \multicolumn{4}{c|}{\textbf{ANAFI~~~~~~~}}                                                                                       \\
\multicolumn{2}{|c||}{\textbf{~}}                                                                                           & \textbf{4K}                   & \textbf{Thermal}              & \textbf{USA} 	    & \textbf{Ai}               \\ \hline \hline
\multirow{9}{*}{\rotatebox{90}{\textbf{Drone~~~~~~~~~~~~~~~}}}      & \textbf{Size folded}\specialref{size} [\si{mm}]        & $244 \times 67 \times 65$     & $218 \times 69 \times 64$     & $252 \times 104 \times 82$    & $304 \times 130 \times 118$   \\ \cline{2-6}
                                                                    & \textbf{Size unfolded}\specialref{size} [\si{mm}] 	    & $242 \times 315 \times 65$    & $242 \times 315 \times 64$    & $303 \times 398 \times 84$    & $378 \times 498 \times 118$   \\ \cline{2-6}
                                                                    & \textbf{Weight} [\si{g}] 			                    & $320$                         & $315$                         & $499$             & $898$                     \\ \cline{2-6}
                                                                    & \textbf{Maximum horizontal speed} [\si{m/s}] 			& \multicolumn{2}{c|}{$15$}                                     & $14.7$            & $16$                      \\ \cline{2-6}
                                                                    & \textbf{Maximum vertical speed} [\si{m/s}] 			& \multicolumn{4}{c|}{$4$}                                                                                      \\ \cline{2-6}
                                                                    & \textbf{Maximum wind resistance} [\si{m/s}] 			& \multicolumn{2}{c|}{$13.9$}                                     & $14.7$           & $12.7$                   \\ \cline{2-6}
                                                                    & \textbf{Service ceiling} [\si{m}]   & \multicolumn{2}{c|}{$4500$}                           & $6000$            & $5000$                    \\ \cline{2-6}
                                                                    & \textbf{Operating temperatures} [\si{\degree C}]    & \multicolumn{2}{c|}{$-10$ to $+40$}                           & $-35$ to $+43$    & $-10$ to $+40$            \\ \cline{2-6}
                                                                    & \textbf{Ingress protection}                           & \multicolumn{2}{c|}{\xmark}                                   & IP53\specialref{ip53}     & IPX3\specialref{ipx3}          \\ \cline{2-6}
                                                                    & \textbf{Noise emission}\specialref{ne} [\si{dB}]     & $64$                & $66$                        & $79$              & $82$                      \\ \cline{2-6}
                                                                    & \textbf{Slots}                                        & \multicolumn{3}{c|}{MicroSD}                                                      & MicroSD \& SIM card       \\ \hline \hline
\multicolumn{2}{|l||}{\textbf{Satellite navigation}}                                                                        & \multicolumn{2}{c|}{GPS \& Glonass}                           &  \multicolumn{2}{c|}{GPS, Glonass \& Galileo} \\ \hline \hline
\multirow{5}{*}{\rotatebox{90}{\textbf{EO Camera}}}                 & \textbf{Sensor}                                       & \multicolumn{4}{c|}{CMOS}                                                                                     \\ \cline{2-6}
                                                                    & \textbf{Aperture}                                     & \multicolumn{3}{c|}{f$/2.4$}                                                      & f$/2.0$                   \\ \cline{2-6}
                                                                    & \textbf{ISO}                                          & \multicolumn{3}{c|}{$100 - 3200$}                                                 & $50 - 6400$               \\ \cline{2-6}
                                                                    & \textbf{Shutter speed} [\si{s}]                       & \multicolumn{3}{c|}{$1 - 1/10000$}                                                & $1/15 - 1/10000$          \\ \cline{2-6}
                                                                    & \textbf{Zoom} [\si{x}]                                & \multicolumn{2}{c|}{$1 - 3$}                                  & $1 - 32$          & $1 - 6$                   \\ \hline
\multirow{5}{*}{\rotatebox{90}{\textbf{Video}}}                     & \textbf{Format}                                       & \multicolumn{3}{c|}{MP4 (H.264)}                                                  & MP4 (H.264, H.265)          \\ \cline{2-6}
                                                                    & \textbf{Resolution}                                   & 4K \& FHD                     & \multicolumn{2}{c|}{4K, FHD \& HD}                & 4K \& FHD                 \\ \cline{2-6}
                                                                    & \textbf{Framerate} [\si{fps}]                         & $24 - 60$                     & $24 - 120$                    & $24 - 30$         & $24 - 120$                \\ \cline{2-6}
                                                                    & \textbf{Horizontal field of view} [\si{\degree}]        & \multicolumn{3}{c|}{$69$}                                                         & $68$                      \\ \cline{2-6}
                                                                    & \textbf{Maximum video bandwidth} [\si{Mbps}]          & \multicolumn{2}{c|}{$100$}                                    & $5$               & $200$                     \\ \hline
\multirow{3}{*}{\rotatebox{90}{\textbf{Photo}}}                   & \textbf{Format}                                       & \multicolumn{4}{c|}{JPEG \& DNG (RAW)}                                                                        \\ \cline{2-6}
                                                                    & \textbf{Resolution} [\si{MP}]                         & \multicolumn{3}{c|}{$21$}                                                         & $48$                      \\ \cline{2-6}
                                                                    & \textbf{Horizontal field of view} [\si{\degree}]        & \multicolumn{2}{c|}{$84$}                                     & $75$              & $73$                      \\ \hline \hline
\multirow{9}{*}{\rotatebox{90}{\textbf{Thermal Camera}}}            & \textbf{Sensor}                                       & \multirow{9}{*}{\xmark}       & FLIR LEPTON 3.5       & FLIR BOSON  & \multirow{9}{*}{\xmark}     \\ \cline{2-2}\cline{4-5}
                                                                    & \textbf{Resolution} [\si{pixels}]                     &                               & $160 \times 120$  & $320 \times 256$  &                           \\ \cline{2-2}\cline{4-5}
                                                                    & \textbf{Temperature range} [\si{\degree C}]         &                               & $-10$ to $+400$               & $-40$ to $+180$   &               \\ \cline{2-2}\cline{4-5}
                                                                    & \textbf{Thermal sensitivity} [\si{\degree C}]       &                               & \multicolumn{2}{c|}{$0.05$}                       &               \\ \cline{2-2}\cline{4-5}
                                                                    & \textbf{Pixel pitch} [\si{\mu m}]                     &                               & \multicolumn{2}{c|}{$12$}                         &               \\ \cline{2-2}\cline{4-5}
                                                                    & \textbf{Horizontal field of view} [\si{\degree}]        &                               & $57$                          & $50$              &               \\ \cline{2-2}\cline{4-5}
                                                                    & \textbf{Photo format}                                 &                               & \multicolumn{2}{c|}{JPEG}                         &               \\ \cline{2-2}\cline{4-5}
                                                                    & \textbf{Video format}                                 &                               & \multicolumn{2}{c|}{MP4 (H.264)}                  &               \\ \cline{2-2}\cline{4-5}
                                                                    & \textbf{Video framerate} [\si{fps}]                   &                               & \multicolumn{2}{c|}{$9$}                          &                           \\ \hline \hline
\multirow{3}{*}{\rotatebox{90}{\textbf{Gimbal}}}                    & \textbf{Mechanical}                                   & \multicolumn{3}{c|}{$2$-axis (roll, pitch)}                                   & $3$-axis   \\ \cline{2-6}
                                                                    & \textbf{Electronic (EIS)}                             & \multicolumn{4}{c|}{$3$-axis}                                                              \\ \cline{2-6}
                                                                    & \textbf{Tilt range}\specialref{tr} [\si{\degree}]    & \multicolumn{3}{c|}{from $-90$ to $+90$}                   & from $-116$ to $+176$          \\ \hline \hline
\multirow{6}{*}{\rotatebox{90}{\textbf{Battery}}}                   & \textbf{Maximum flight time} [\si{min}] 			    & $25$                          & $26$                          & \multicolumn{2}{c|}{$32$}                     \\ \cline{2-6}
                                                                    & \textbf{Type}                                         & \multicolumn{2}{c|}{LiPo (2 cells)}                           & \multicolumn{2}{c|}{LiPo (3 cells)}           \\ \cline{2-6}
                                                                    & \textbf{Capacity} [\si{mAh}]                          & \multicolumn{2}{c|}{$2700$}                                   & $3400$            & $6800$                    \\ \cline{2-6}
                                                                    & \textbf{Voltage} [\si{V}]                             & \multicolumn{2}{c|}{$7.6$}                                    & \multicolumn{2}{c|}{$13.2$}                    \\ \cline{2-6}
                                                                    & \textbf{Weight} [\si{g}]                              & \multicolumn{2}{c|}{$125$}                                    & $195$             & $366$                     \\ \cline{2-6}
                                                                    & \textbf{Maximum charging power} [\si{W}]              & \multicolumn{2}{c|}{$25$}                                     & $30$              & $45$                      \\ \cline{2-6}
                                                                    & \textbf{Charging port}                                & \multicolumn{4}{c|}{USB-C}                                                                                    \\ \hline \hline
\multicolumn{2}{|l||}{\textbf{Controller}}                                                                                  & \multicolumn{3}{c|}{\nameref{subsubsec:skycontroller3}}               & \nameref{subsubsec:skycontroller4}    \\ \hline \hline
\multirow{3}{*}{\rotatebox{90}{\textbf{Tools}}}                       & \href{https://developer.parrot.com/docs/airsdk/index.html}{\textbf{Air SDK}}                                      & \xmark                        & \xmark                        & \xmark            & \cmark                    \\ \cline{2-6}
                                                                    & \href{https://developer.parrot.com/docs/sphinx/index.html}{\textbf{Sphinx}}                                       & \cmark                        & \xmark                        & \xmark            & \cmark                    \\ \cline{2-6}
                                                                     & \cellcolor{yellow!20}\href{https://github.com/andriyukr/anafi_ros}{\textbf{anafi\_ros}}                                        & \cellcolor{yellow!20}\cmark                        & \cellcolor{yellow!20}\cmark                        & \cellcolor{yellow!20}\cmark            & \cellcolor{yellow!20}\cmark                    \\ \hline
\end{tabular}
\label{tab:anafi}
\end{table*}

\begin{table}[!b]
\vspace{-0.5cm}
\setlength\tabcolsep{0pt}
\begin{tabular}{m{8.6cm}}
\mbox{~~~\speciallabel{size}}\footnotesize{length $\times$ width $\times$ height} \\ 
\mbox{~~~\speciallabel{ip53}}\footnotesize{protected from limited dust ingress and water spray less than $60^{\circ}$ from vertical} \\
~~~\speciallabel{ipx3}\footnotesize{protected from water spray less than $60^{\circ}$ from vertical} \\ 
~~~\speciallabel{ne}\footnotesize{at $1 ~\si{m}$} \\
~~~\speciallabel{tr}\footnotesize{from nadir to zenith}
\end{tabular}
\end{table}

\begin{table*}[!t]
\caption{Technical specifications of Parrot Skycontrollers.}
\centering
\begin{tabular}{|l||c|c|} \hline
\multirow{2}{*}{\textbf{Parameter}}     & \multicolumn{2}{c|}{\textbf{Skycontroller~~~~~~~~~~~~~~~}}        \\
                                        & \textbf{3}                    & \textbf{4}                        \\ \hline \hline
\textbf{Size folded}\specialref{size} [\si{mm}]          & $94 \times 152 \times 72$     & $147 \times 238 \times 55$        \\ \hline
\textbf{Size unfolded}\specialref{size} [\si{mm}]        & $153 \times 152 \times 116$   & $147 \times 315 \times 55$        \\ \hline
\textbf{Weight [\si{g}]}                & $386$                         & $606$                             \\ \hline
\textbf{Transmission system}            & Wi-Fi 802.11a/b/g/n           & Wi-Fi 802.11a/b/g/n \& 4G         \\ \hline
\textbf{Frequency used [\si{GHz}]}      & $2.4$, $5.8$                  & $2.4$, $5$                        \\ \hline
\textbf{Maximum transmission distance [\si{km}]}    & $4$               & $\infty$                          \\ \hline
\textbf{Video stream resolution}        & HD $720$p                     & $1080$p                           \\ \hline
\textbf{Battery capacity [\si{mAh}]}    & $2500$                        & $3350$                            \\ \hline
\textbf{Battery life [\si{h}]}          & \multicolumn{2}{c|}{$2.5$~~~~~~~~~~~~~~~}                         \\ \hline
\textbf{Ports}  & USB-C (charge) \& USB-A (connection) &    USB-C (charge and connection) \& micro-HDMI     \\ \hline
\textbf{Compatible mobile devices}      & screen size up to $6$''       & screen size up to $8$''           \\ \hline
\textbf{Ingress protection}             & \xmark                        & IP5X\specialref{ip5x}             \\ \hline
\textbf{Compatible drones}              & \nameref{subsubsec:anafi_4k}, \nameref{subsubsec:anafi_thermal}, \nameref{subsubsec:anafi_usa}  & \nameref{subsubsec:anafi_ai}    \\ \hline
\rowcolor{yellow!20} \textbf{Support in \href{https://github.com/andriyukr/anafi_ros}{anafi\_ros}}  & \cmark    & \cmark            \\ \hline
\end{tabular}
\label{tab:skycontroller}
\end{table*}

\begin{table}[!b]
\vspace{-0.5cm}
\setlength\tabcolsep{0pt}
\begin{tabular}{m{8.6cm}}
\mbox{~~~\speciallabel{ip5x}}\footnotesize{protected from limited dust ingress}
\end{tabular}
\end{table}

\subsubsection{ANAFI 4K}
\label{subsubsec:anafi_4k}

This model (shown in Fig.~\ref{fig:anafi_4k}) is the first and basic drone of the ANAFI family. ANAFI 4K is one of the quietest drones in its class with a noise level of $64~\si{dB}$ at $1~\si{m}$. With $25~\si{min}$ of flight time, the battery can be recharged via a USB-C cable in $90~\si{min}$. The model of ANAFI~4K is available in Parrot's software-in-the-loop simulation environment -- Sphinx.

\subsubsection{ANAFI Thermal}
\label{subsubsec:anafi_thermal}

This model (shown in Fig.~\ref{fig:anafi_thermal}) is the upgrade of ANAFI 4K with a thermal camera. The optical unit of ANAFI Thermal combines the electro-optics with an infrared sensor, making it possible to identify temperatures between $-10~\si{\degree C}$ and $+400~\si{\degree C}$. Thanks to the FLIR Lepton radiometric sensor, the absolute temperature of each pixel can be determined. The RGB image can be blended with thermal images. This enables to detection of hot spots with the thermal camera, while the RGB camera allows the viewing of important details. Despite the thermal camera, ANAFI Thermal is the smallest and lightest model of the family.

\subsubsection{ANAFI USA}
\label{subsubsec:anafi_usa}

This model (shown in Fig.~\ref{fig:anafi_usa}) is the rescue-grade drone, featuring $32\si{x}$ zoom and thermal imaging capabilities to meet the demands of first responders and search-and-rescue teams. To achieve this, ANAFI~USA is equipped with three front-mounted cameras: a thermal camera, $21~\si{Mpx}$ RGB wide-angle camera (for $1\si{x}$ to $5\si{x}$ zoom) and $21~\si{Mpx}$ RGB telephoto camera (for $5\si{x}$ to $32\si{x}$ zoom), which guarantees a continuous zoom. The $32\si{x}$ zoom allows seeing details as small as $1~\si{cm}$ from a distance of $50~\si{m}$. The image stabilisation system of ANAFI~USA ensures high-quality footage even at $15~\si{m/s}$ wind gust. Despite its compact design, ANAFI~USA boasts a $32~\si{min}$ of flight time. ANAFI~USA has IP53\specialref{ip53} ingress protection, offering water and dust resistance and making it suitable to fly in rainy conditions. ANAFI~USA has a service ceiling of $6~\si{km}$ and can operate in temperatures between $-35~\si{\degree C}$ and $+43~\si{\degree C}$. The body of ANAFI~USA is mainly made of polyamide, reinforced with carbon fibre and streamlined using hollow glass beads. The data stored on ANAFI~USA or sent through the networks are encrypted, and the drone is protected against malicious software modification attempts.

\subsubsection{ANAFI Ai}
\label{subsubsec:anafi_ai}

This model (shown in Fig.~\ref{fig:anafi_ai}) is the biggest and heaviest but most advanced of the ANAFI family. Anafi~Ai is the first drone to use the 4G cellular network connection, in addition to Wi-Fi, as an alternative encrypted data link between the drone and the remote controller, theoretically enabling control at any distance.
Besides a high-resolution $48~\si{Mpx}$ RGB camera with an ISO range of $50 - 6400$, ANAFI~Ai is also equipped with a pair of multidirectional stereo cameras, which allow the computation of the occupancy grid to avoid obstacles automatically. Anafi~Ai has a $3$-axis gimbal differently from other Anafi models with $2$-axis gimbals.
The maximum horizontal speed of Anafi~Ai is $16~\si{m/s}$, thanks to the optimized aerodynamic performance of the vehicle. Anafi~Ai's $6800~\si{mAh}$ battery allows $32~\si{min}$ of flight time and can be recharged in $150~\si{min}$.
ANAFI~Ai has IPX3\specialref{ipx3} ingress protection, offering water resistance and making it suitable to fly in rainy conditions.
ANAFI~Ai can execute custom C++ and Python code onboard, thanks to Parrot's Air SDK. Air SDK allows loading and running code directly on ANAFI~Ai and accessing all sensors, connectivity interfaces and autopilot features.  
The model of ANAFI~Ai is available in the Sphinx simulation environment.

\subsection{Remote Controllers}

\begin{figure}[!b]
\centering
\subfloat[Skycontroller 3.]{\includegraphics[width=0.35\columnwidth]{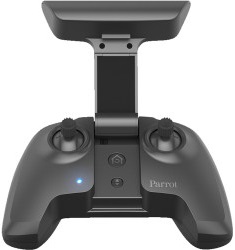}%
\label{fig:skycontroller_3}} \hfill
\subfloat[Skycontroller 4.]{\includegraphics[width=0.55\columnwidth]{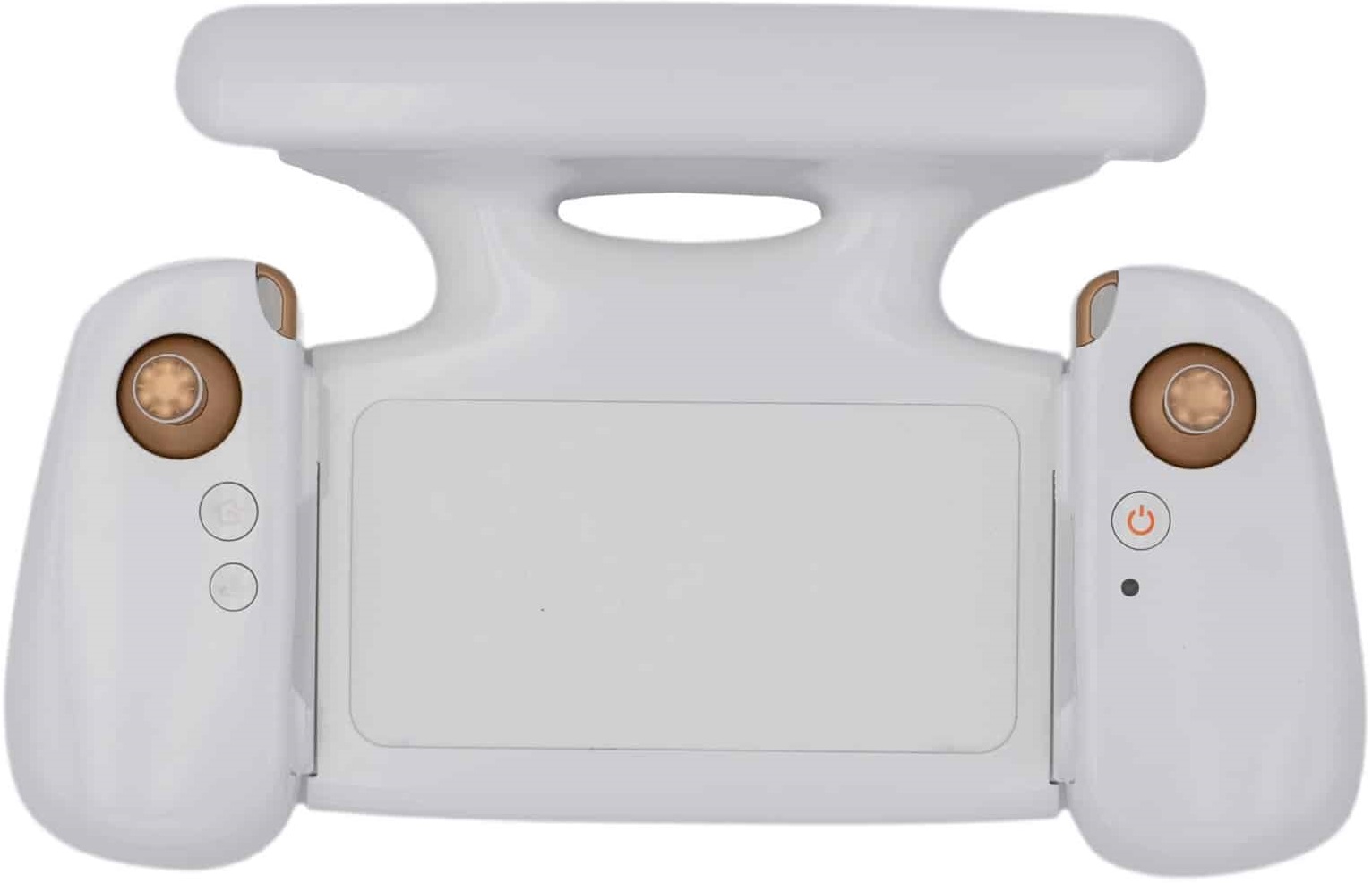}%
\label{fig:skycontroller_4}}
\caption{Parrot Skycontroller series remote controllers.}
\label{fig:skycontroller}
\end{figure}

Parrot ANAFI drones come with a handheld remote radio controller called Skycontroller, allowing the user to control the drone and access its various features. Skycontrollers feature a light and compact design with a conventional layout of sticks and buttons. Moreover, they have a built-in battery rechargeable through the \mbox{USB-C} port. The Parrot ANAFI family has two Skycontroller models: \textit{Skycontroller~3} for ANAFI~4K, ANAFI~Thermal and ANAFI~USA, and \textit{Skycontroller~4} for ANAFI~Ai. The two versions of Skycontrollers are illustrated in Fig.~\ref{fig:skycontroller}, while Table~\ref{tab:skycontroller} summarises the properties of each model.

\newcommand{\TwoFiguresWidthTwo}{0.45}

\subsubsection{Skycontroller 3}
\label{subsubsec:skycontroller3}

This model (shown in Fig.~\ref{fig:skycontroller_3}) is a remote radio controller designed for Parrot ANAFI 4K, Thermal and USA drones. Skycontroller~3 has a maximum range of up to $4~\si{km}$. 

\subsubsection{Skycontroller 4}
\label{subsubsec:skycontroller4}

This model (shown in Fig.~\ref{fig:skycontroller_4}) is a remote 4G controller designed for Parrot ANAFI~Ai. Skycontroller~4 has IP5X\specialref{ip5x} ingress protection, offering dust resistance.

\section{Developed Framework}
\label{sec:framework}

The developed framework -- {\tt anafi\_ros} -- is a python-based ROS package which enables interfacing with all available Parrot ANAFI quadrocopters. Besides being compatible with all physical drones, {\tt anafi\_ros} can connect to virtual drones in Parrot's simulation environment -- Sphinx. The developed {\tt anafi\_ros} is built on top of Parrot’s official python SDK -- Olympe -- which provides a programming interface for Parrot ANAFI drones. The communication flow in {\tt anafi\_ros} allows connecting directly to the drones via Wi-Fi interfaces or through Skycontrollers via USB ports, which is highly recommended. The developed framework makes connecting multiple drones to the same ground station easy by automatically assigning a different virtual IP address to each connected Skycontroller and managing port forwarding. The main functionalities of {\tt anafi\_ros} include drone piloting, feedback of flight parameters from onboard sensors, gimbal control, drone state monitoring, video streaming from onboard cameras, picture capturing, video recording, file transferring between onboard storage and ground station, drone calibration and flight plan management.

\begin{remark}
For the complete list of subscribed and published topics, available services and parameters, please refer to \hyperref[sec:appendix]{Appendix}.
\end{remark}

\begin{figure}[!b]
\centering
    \resizebox{0.99\columnwidth}{!}{
    \begin{tikzpicture}
        \begin{umlpackage}[name=anafi-ros]{anafi\_ros}
            \umlclass[x=0, y=0]{skycontroller}{
                drone : Drone
            }{
                connect(ip : string) : void \\
                run() : void
            }
            \umlclass[below=3cm of skycontroller]{anafi}{
                drone : Drone
            }{
                connect(ip : string) : void \\
                \umlvirt{run() : void}
            }
            \umlclass[below left=1cm and 1.8cm of anafi]{4k}{}{
                run() : void
            }
            \umlclass[right=0.1cm of 4k]{thermal}{}{
                run() : void
            }
            \umlclass[right=0.1cm of thermal]{usa}{}{
                run() : void
            }
            \umlclass[right=0.1cm of usa]{ai}{}{
                run() : void
            }
            \umlclass[below left=-1.2cm and 1cm of anafi, width=2.5cm]{storage}{}{}
            \umlclass[above=0.5cm of storage, width=2.5cm]{gimbal}{}{
                rotate() : void
            }
            \umlclass[above=0.5cm of gimbal, width=2.5cm]{battery}{}{}
            \umlclass[below right=1cm and -1.4 of thermal]{camera}{}{
                stream() : Image
            }        
            \umlNarynode[y=-2.5, name=connection]{} \node at (-1.3,-2.5) {\small connection};
            \umlclass[right=1cm of connection]{link}{}{}
        \end{umlpackage}
        \umlactor[above right=-1cm and 2.15cm of skycontroller, below=-1cm]{operator}
        \begin{scope}[shift={(4.5, -5.2)}, scale=0.2]
            \draw (-1,-1) -- (-1,1) -- (1,1) -- (1,-1) -- (-1,-1) -- (-2,-2) -- (2,-2) -- (1,-1);
            \node at (0,-3) {\small computer};
        \end{scope}
        
        \umlassoc[geometry=--, mult2=1]{connection}{anafi}
        \umlassoc[geometry=--, mult1=1]{skycontroller}{connection}
        \umlassoc[geometry=--, mult2=1]{connection}{link}
        \umlinherit[geometry=|-|]{4k}{anafi}
        \umlinherit[geometry=|-|]{thermal}{anafi}
        \umlinherit[geometry=|-|]{usa}{anafi}
        \umlinherit[geometry=|-|]{ai}{anafi}

        \draw[-] (battery.east) -| node[pos=0.1, below]{\small 1} ($(anafi.west) + (-0.3cm, 0.5cm)$) -- node[pos=0.4, below]{\small 1} ($(anafi.west) + (0, 0.5cm)$);
        \draw[-] (gimbal.east) -| node[pos=0.2, below]{\small 1} ($(anafi.west) + (-0.7cm, 0)$) -- node[pos=0.7, below]{\small 1} (anafi.west);
        \draw[-] (storage.east) -- node[pos=0.3, below]{\small 0..1} node[pos=0.8, below]{\small 1} ($(storage.east) + (1cm, 0)$);
        \draw[-] (camera.west) -| node[pos=0.1, below]{\small 1} node[pos=0.9, left]{\small 1} (4k.south);
        \draw[-] ($(camera.north) + (-1cm, 0)$) -- node[pos=0.2, left]{\small 2} node[pos=0.8, left]{\small 1} ($(camera.north) + (-1cm, 1cm)$);
        \draw[-] ($(camera.north) + (1cm, 0)$) -- node[pos=0.2, left]{\small 3} node[pos=0.8, left]{\small 1} ($(camera.north) + (1cm, 1cm)$);
        \draw[-] (camera.east) -| node[pos=0.1, above]{\small 4} node[pos=0.9, left]{\small 1} (ai.south);
        \draw[-] ($(operator.west) + (-2cm, 0)$) -- node[pos=0.2, below]{\small 0..*} node[pos=0.8, below]{\small 0..1} (operator.west);
        \draw[-] ($(skycontroller.east) + (0, -0.9cm)$) -| node[pos=0.09, below]{\small 0..*} node[pos=0.97, left]{\small 0..1} ($(operator.south) + (-0.2cm, -4.5cm)$);
        \draw[-] (anafi.east) -- node[pos=0.2, below]{\small 0..*} node[pos=0.8, below]{\small 0..1} ($(anafi.east) + (2cm, 0)$);
        \draw[-] ($(operator.south) + (0.2cm, 0)$) -- node[pos=0.05, right]{\small 1..*} node[pos=0.95, right]{\small 0..*} ($(operator.south) + (0.2cm, -4.5cm)$);
    \end{tikzpicture}}
\caption{UML diagram of the structure of {\tt anafi\_ros}.}
\label{fig:uml_diagram}
\end{figure}
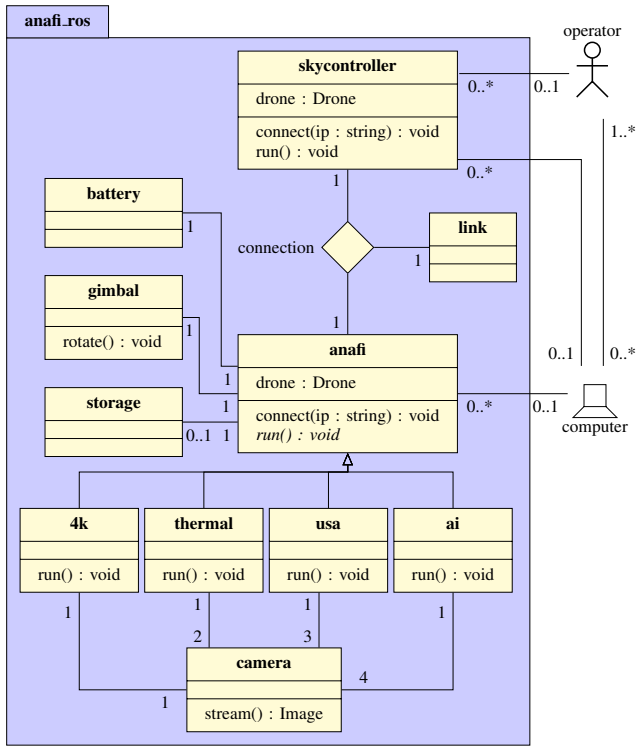

The developed package is organised with several subelements to facilitate the development, as depicted in Fig.~\ref{fig:uml_diagram}. In other words, each physical component, such as the drone itself, gimbal, camera, battery, connection link, storage device and remote controller, has a respective software element.

\subsubsection{Drone}

The drone element is a core part of the package, which manages the connection to the drone, enables the control of the drone and provides feedback from the drone. This element allows piloting the drone in three modes: directly commanding values in~\eqref{eq:u_virtual}, commanding relative displacements $\begin{bmatrix} \Delta x^* & \Delta y^* & \Delta z^* & \Delta \psi^* \end{bmatrix}^T$ or commanding world references $\begin{bmatrix} \lambda_x^* & \lambda_y^* & z^* & \psi^* \end{bmatrix}^T$, where $\lambda_x^*$ and $\lambda_y^*$ are the desired latitude and longitude, respectively. The drone element retrieves and publishes real-time information, such as the drone's attitude, altitude, speed and GPS location. It forwards to the drone the requests for arming, taking-off and landing. Drone class also allows bounding the altitude, distance, horizontal and vertical speed, pitch and roll angles, and attitude rates.

\subsubsection{Gimbal}

The gimbal element provides control and feedback on the camera's gimbal. This element controls the desired pitch $\phi_G^*$ and roll $\theta_G^*$ of the camera. It also provides the actual attitude $\boldsymbol{\theta}^D_G$ of the gimbal and allows for setting the maximum rotational speed of the gimbal.

\subsubsection{Camera}

The camera element provides essential capabilities for the camera, such as changing the zoom level, capturing pictures and recording videos. This element also publishes real-time video stream, camera calibration matrix and actual zoom level. It also allows setting the camera mode, image style and streaming mode and enabling HDR mode.

\subsubsection{Battery}

The battery element provides the battery status, such as the battery's level, health and voltage.

\subsubsection{Link}

The link element provides information on the connection to the drone, such as the link quality, signal strength and connection throughput.

\subsubsection{Storage}

The storage element provides the available memory on the microSD cards if installed. This element also allows downloading media (photos and videos) from the storage device and formatting it.

\subsubsection{Controller}

The controller element provides an alternative to connect to the drone via the remote controller. This element reads and publishes the state of the sticks (gaz/yaw and pitch/roll), triggers (gimbal tilt and camera zoom) and buttons (return to home, centre camera and reset zoom). The remote controller also streams its real-time attitude.

\subsection{Complementary Packages}

\setcounter{footnote}{0} 
\renewcommand{\thefootnote}{\roman{footnote}}

A complimentary ROS package -- {\tt anafi\_autonomy}\footnote{\url{https://github.com/andriyukr/anafi\_autonomy}}~-- was developed on top of {\tt anafi\_ros} to enable safe navigation of ANAFI drones by adding some high-level capabilities, like position and velocity control.
Besides, other open-source ROS packages are available for building the navigation stack for aerial robots, like visual-inertial localisation~\cite{Mur2015TRo}\footnote{\url{https://github.com/raulmur/ORB_SLAM2}}, environment perception~\cite{Tallamraju2019RAL}\footnote{\url{https://github.com/robot-perception-group/AirPose}}, motion planning~\cite{Zhou2021TRo}\footnote{\url{https://github.com/HKUST-Aerial-Robotics/Fast-Planner}}, model-based
~\cite{Torrente2021RAL}\footnote{\url{https://github.com/uzh-rpg/data_driven_mpc}} and model-free~\cite{Sarabakha2020TFS}\footnote{\url{https://github.com/andriyukr/controllers}} control.


\renewcommand{\TwoFiguresWidth}{0.99}
\begin{figure*}[!b]
\centering
\vspace{-0.5cm}
\subfloat[Pitch tracking.]{\includegraphics[width=\TwoFiguresWidth\columnwidth]{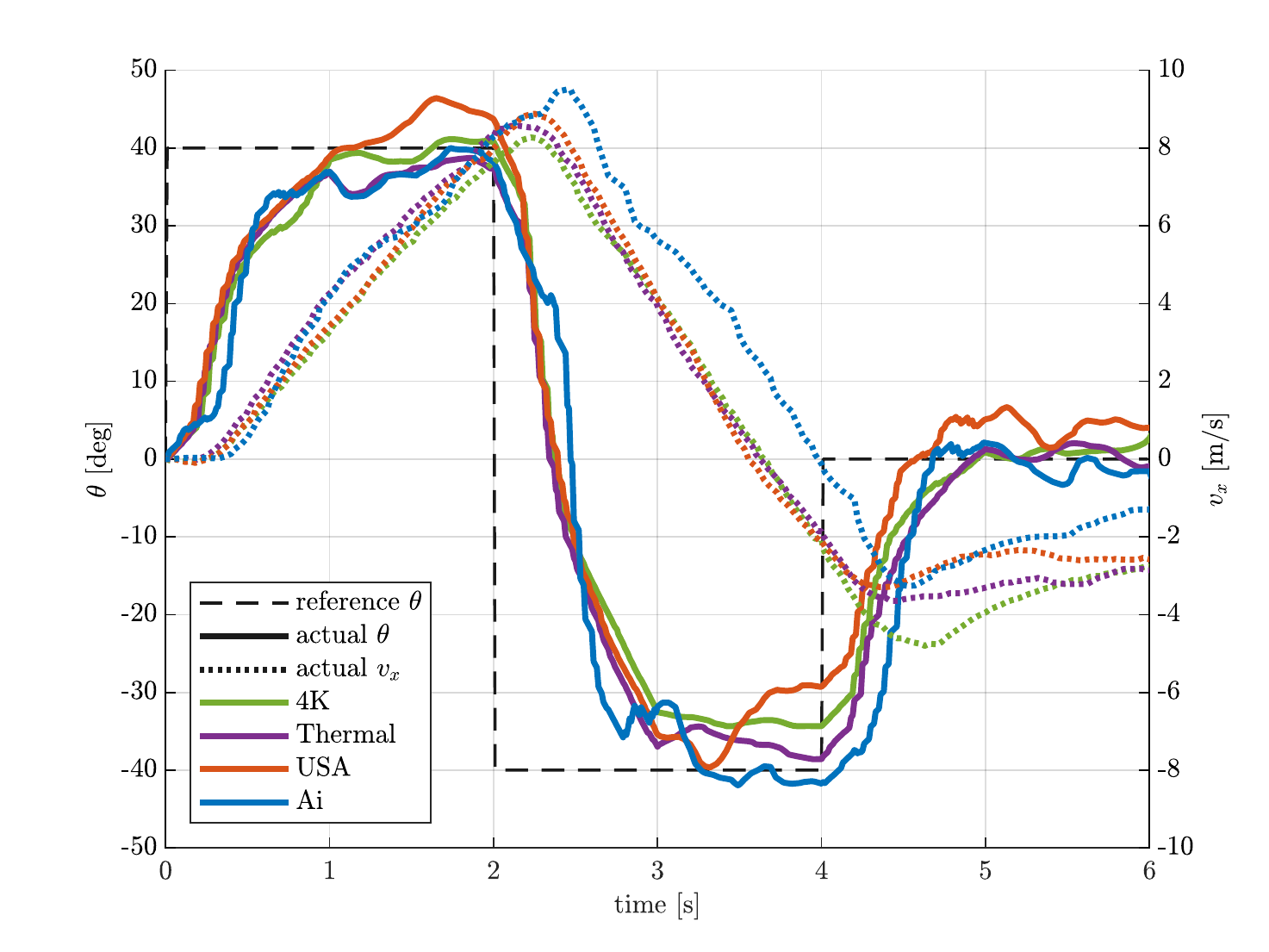}%
\label{fig:piloting_pitch}} \hfill
\subfloat[Roll tracking.]{\includegraphics[width=\TwoFiguresWidth\columnwidth]{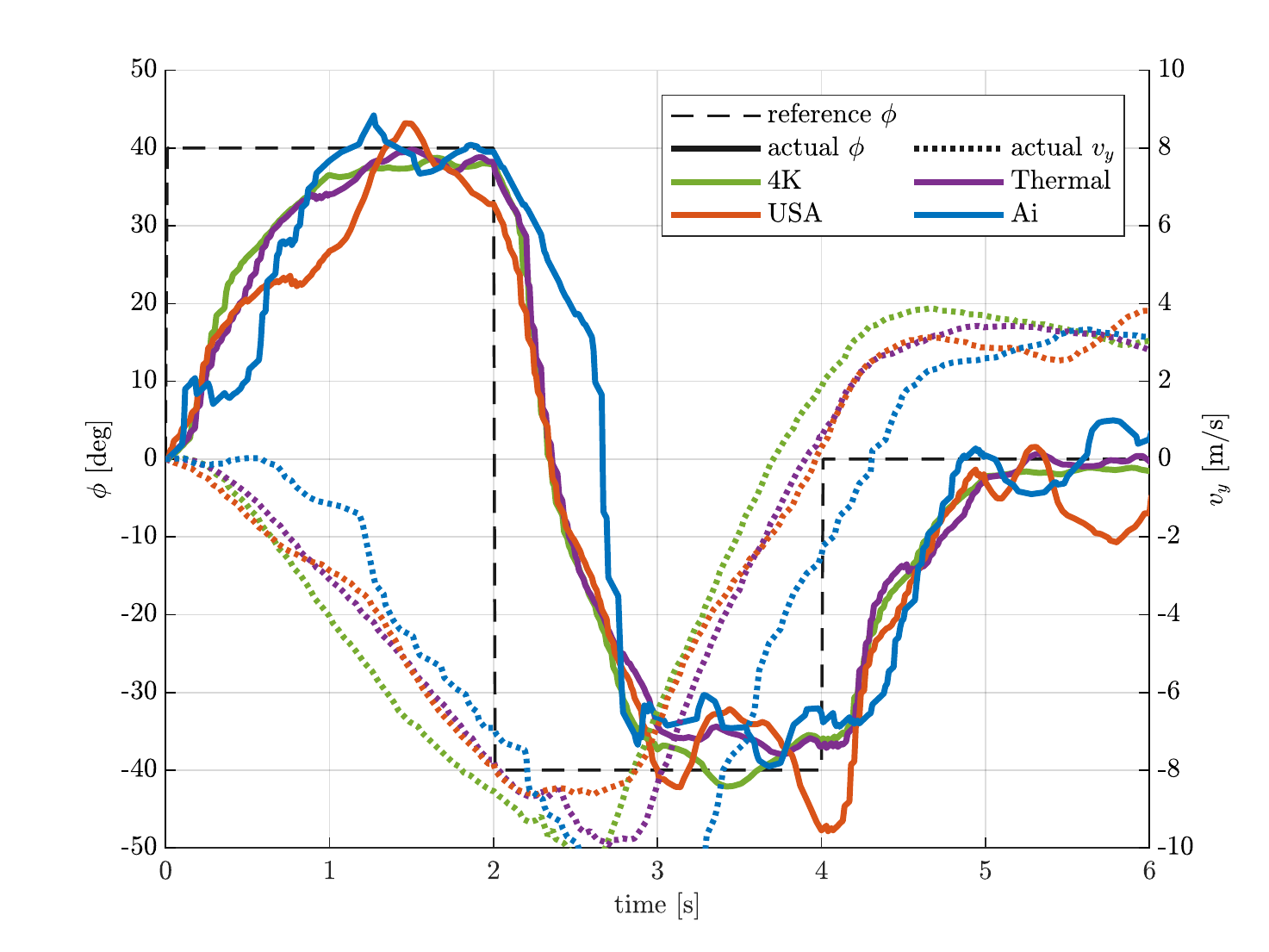}%
\label{fig:piloting_roll}} \\
\subfloat[Vertical speed tracking.]{\includegraphics[width=\TwoFiguresWidth\columnwidth]{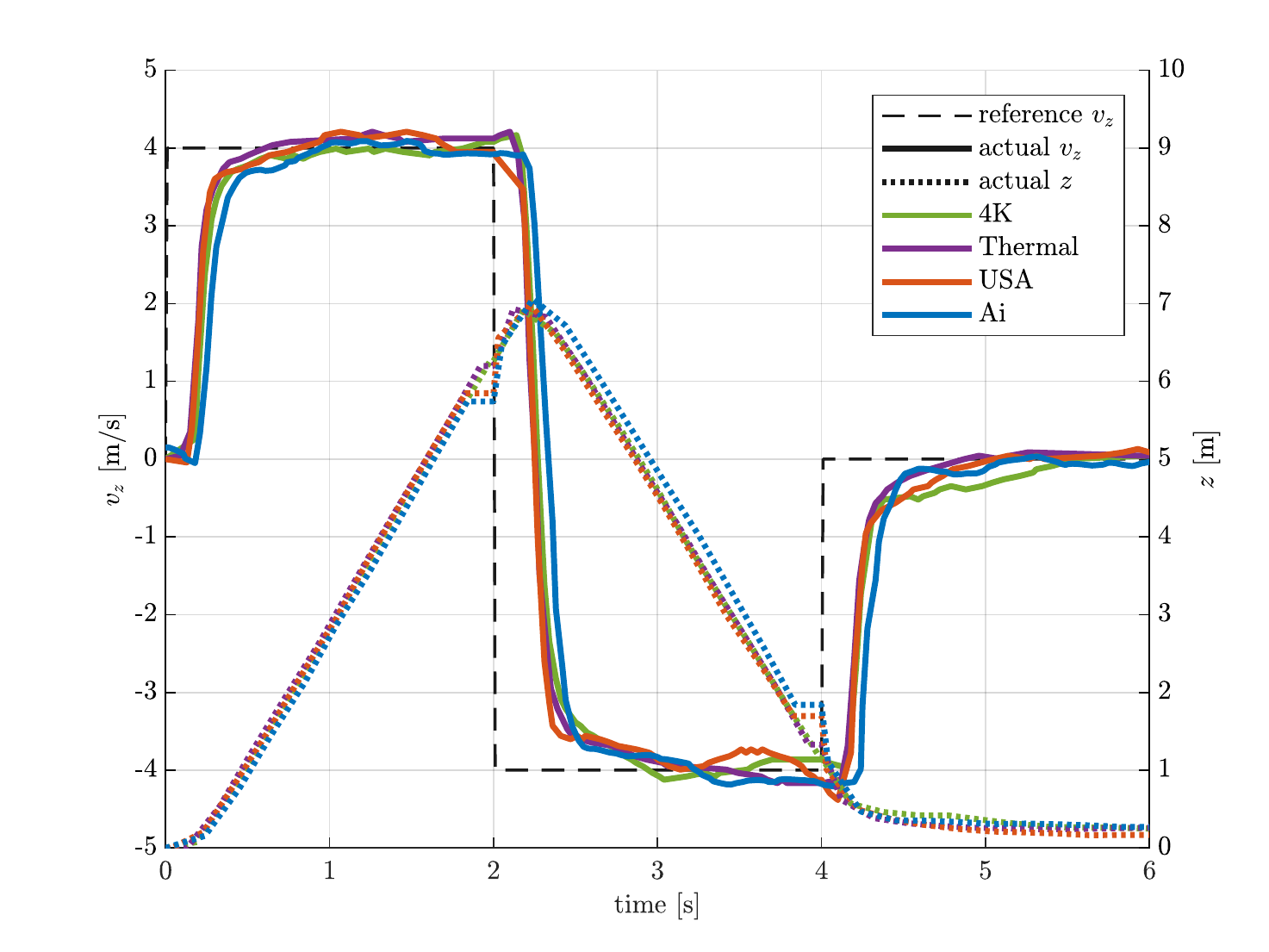}%
\label{fig:piloting_z}} \hfill
\subfloat[Yaw rate tracking.]{\includegraphics[width=\TwoFiguresWidth\columnwidth]{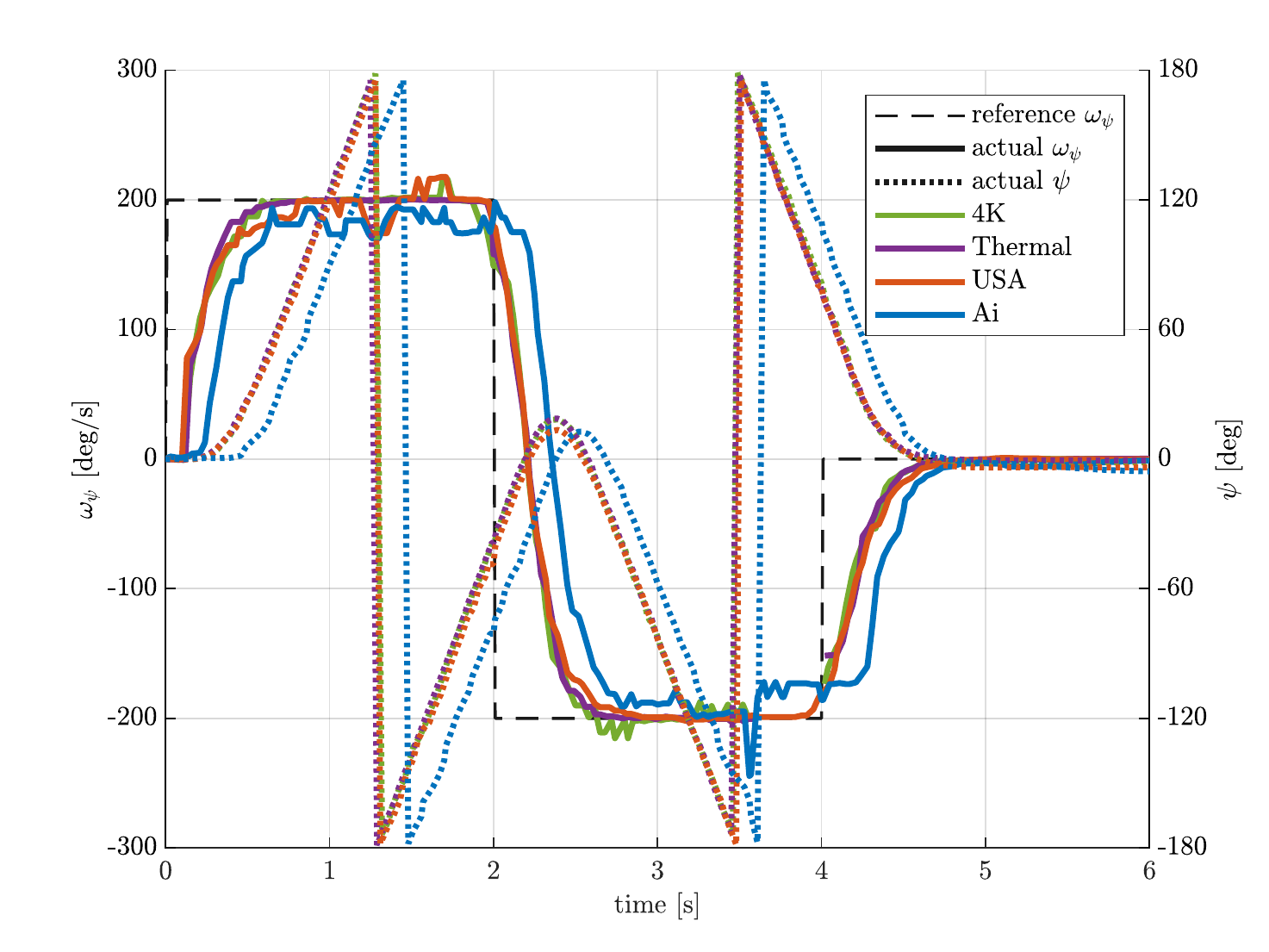}%
\label{fig:piloting_yaw}}
\caption{Piloting response.}
\label{fig:piloting}
\end{figure*}

\section{Experimental Validation}
\label{sec:validation}

To verify the declared characteristics and validate the developed package, we tested the drone's flight characteristics, gimbal response and camera capabilities separately.

\subsection{Drone}

The drones were pushed to their limits by commanding the maximum control inputs to verify the drones' capabilities and validate the developed package. The tests were performed in an open field on a windless day. 

For the pitch tracking response, first, the commanded pitch was set to $40\si{\degree}$ for $2~\si{s}$, then, it was reversed to $-40\si{\degree}$ for $2~\si{s}$ and, finally, set to $0\si{\degree}$. As can be observed from Fig.~\ref{fig:piloting_pitch}, all drones were able to achieve the desired pitch while reaching speeds above $8~\si{m/s}$ and stabilise at around $0\si{\degree}$ in the end. ANAFI~4K and Thermal had smoother behaviour, while ANAFI~USA had a more aggressive response.

Similarly, for the roll tracking response, first, the commanded roll was set to $40\si{\degree}$ for $2~\si{s}$, then, it was reversed to $-40\si{\degree}$ for $2~\si{s}$ and, finally, set to $0\si{\degree}$. As can be observed from Fig.~\ref{fig:piloting_roll}, all drones were able to achieve the desired roll while reaching speeds of above $10~\si{m/s}$ and stabilise at around $0\si{\degree}$ in the end. ANAFI 4K and Thermal still had a stable response, while ANAFI USA and Ai had more twitching behaviour.

For the vertical velocity tracking response, first, the commanded velocity was set to $4~\si{m/s}$ for $2~\si{s}$, then, it was reversed to $-4~\si{m/s}$ for $2~\si{s}$ and, finally, set to $0~\si{m/s}$. As can be observed from Fig.~\ref{fig:piloting_z}, all drones have an initial delay between $100\si{ms}$ and $200\si{ms}$ but later were able to achieve the desired climbing and descend speeds, reaching the altitude of almost $7~\si{m}$ in $2~\si{s}$, and stabilise at around $0~\si{m/s}$ in the end. All drones had smooth and stable behaviour.

For the yaw rate tracking response, first, the commanded yaw rate was set to $200\si{\degree/s}$ for $2~\si{s}$, then, it was reversed to $-200\si{\degree/s}$ for $2~\si{s}$ and, finally, set to $0\si{\degree/s}$. As shown in Fig.~\ref{fig:piloting_yaw}, all drones were able to achieve the desired yaw rate, making a $360\si{\degree}$ turn in less than $2.5~\si{s}$, and stabilise at around $0\si{\degree/s}$ in the end. Similarly, to the velocity tracking, ANAFI 4K, Thermal and USA have a response delay of approximately $100\si{ms}$; while, for ANAFI Ai, it is approximately $200\si{ms}$. After the transient phase, all drones had stable spins at the desired yaw rate. 

\subsection{Gimbal}

\renewcommand{\TwoFiguresWidth}{0.49}
\begin{figure}[!b]
\centering
\subfloat[Roll tracking in orientation mode.]{\includegraphics[width=\TwoFiguresWidth\columnwidth]{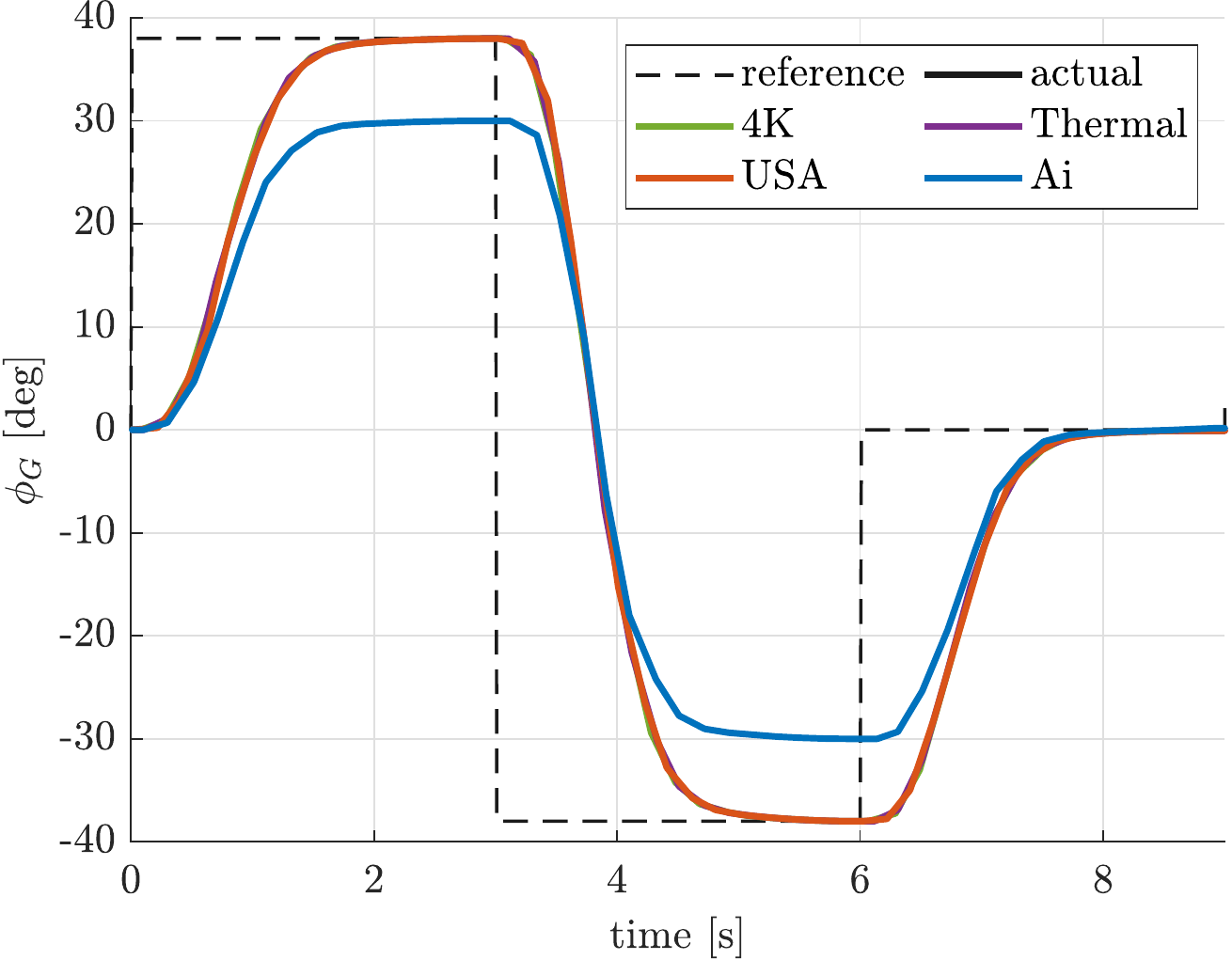}%
\label{fig:gimbal_roll_position}} \hfill
\subfloat[Roll tracking in velocity mode.]{\includegraphics[width=\TwoFiguresWidth\columnwidth]{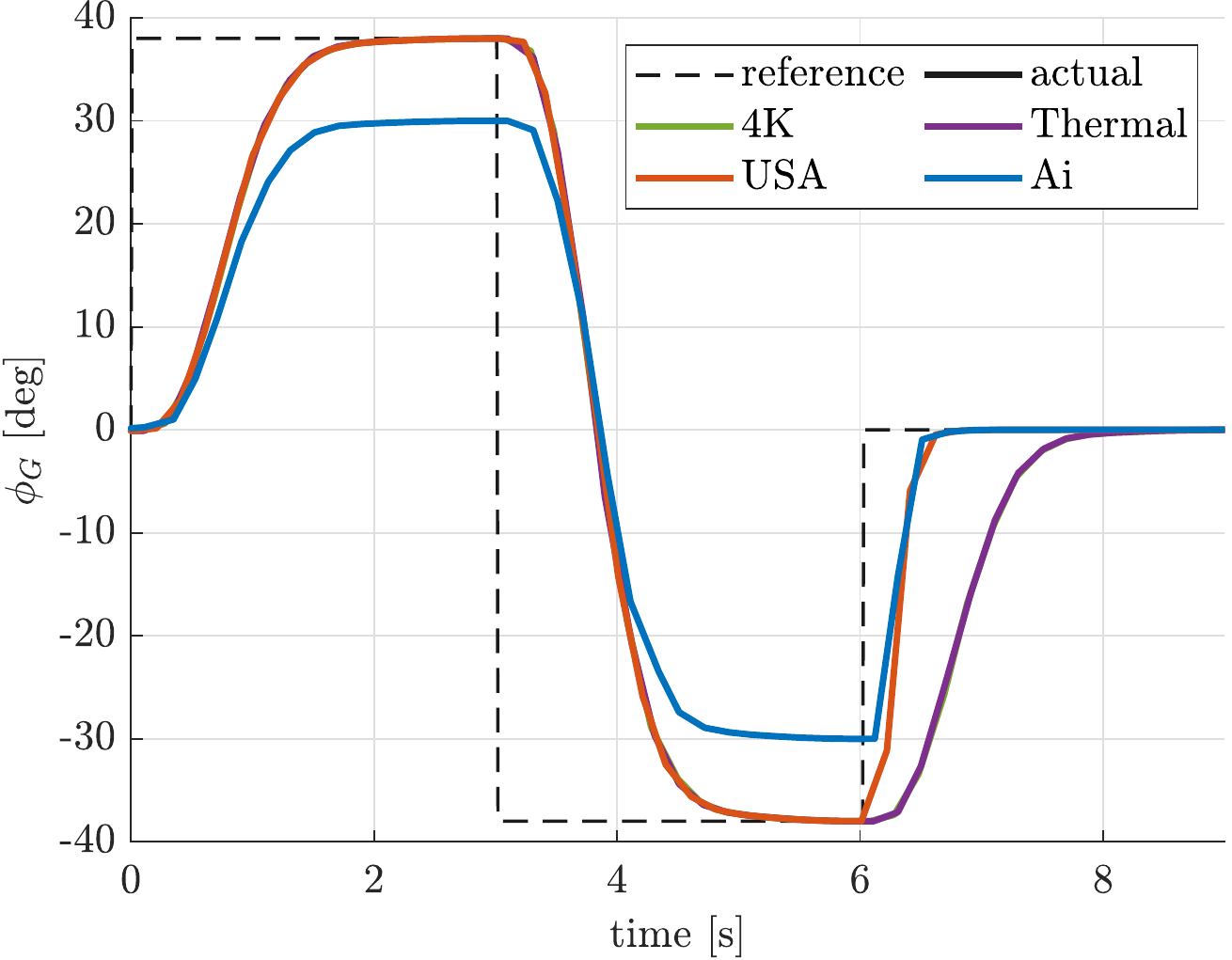}%
\label{fig:gimbal_roll_velocity}} \\
\subfloat[Pitch tracking in orientation mode.]{\includegraphics[width=\TwoFiguresWidth\columnwidth]{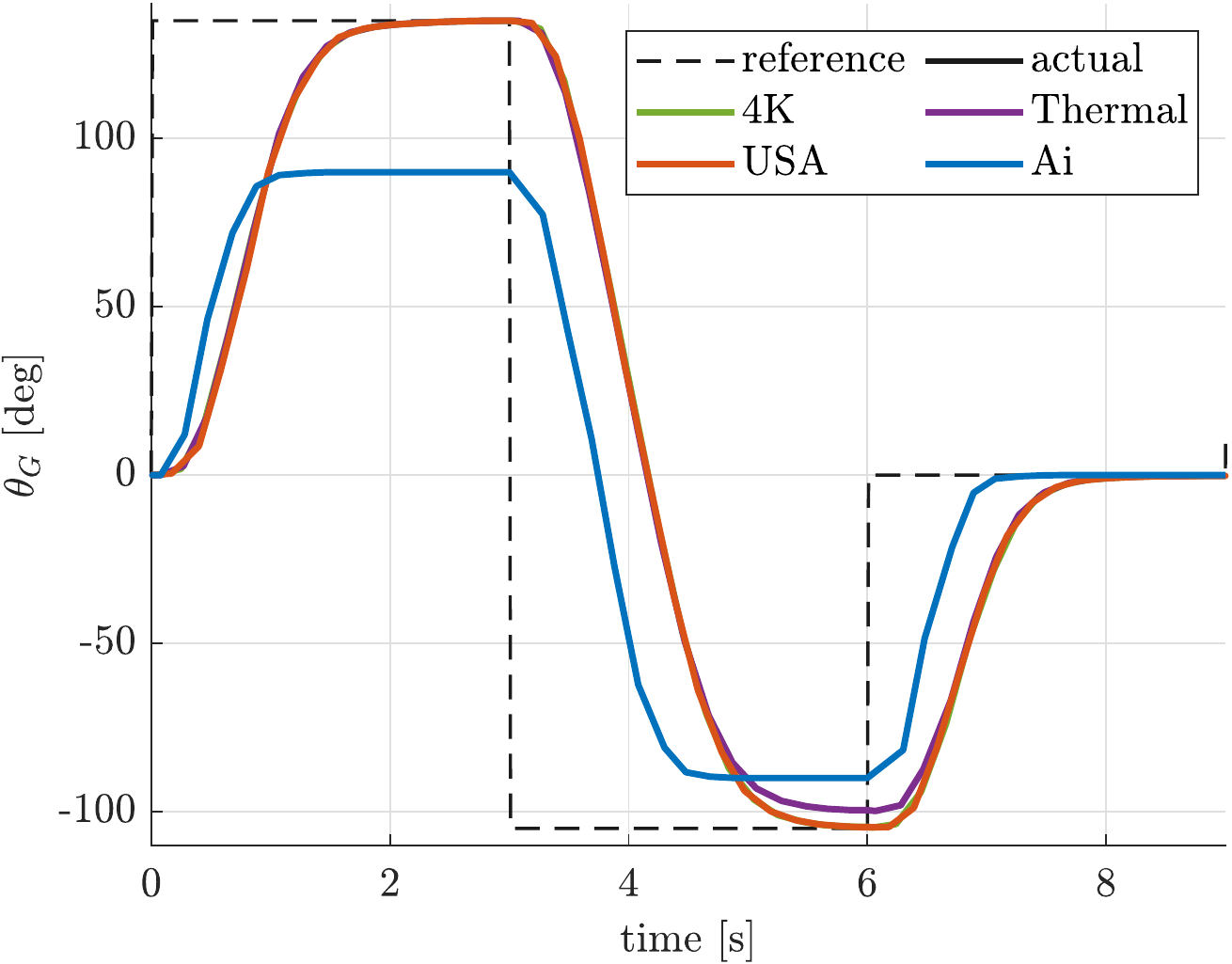}%
\label{fig:gimbal_pitch_position}} \hfill
\subfloat[Pitch tracking in velocity mode.]{\includegraphics[width=\TwoFiguresWidth\columnwidth]{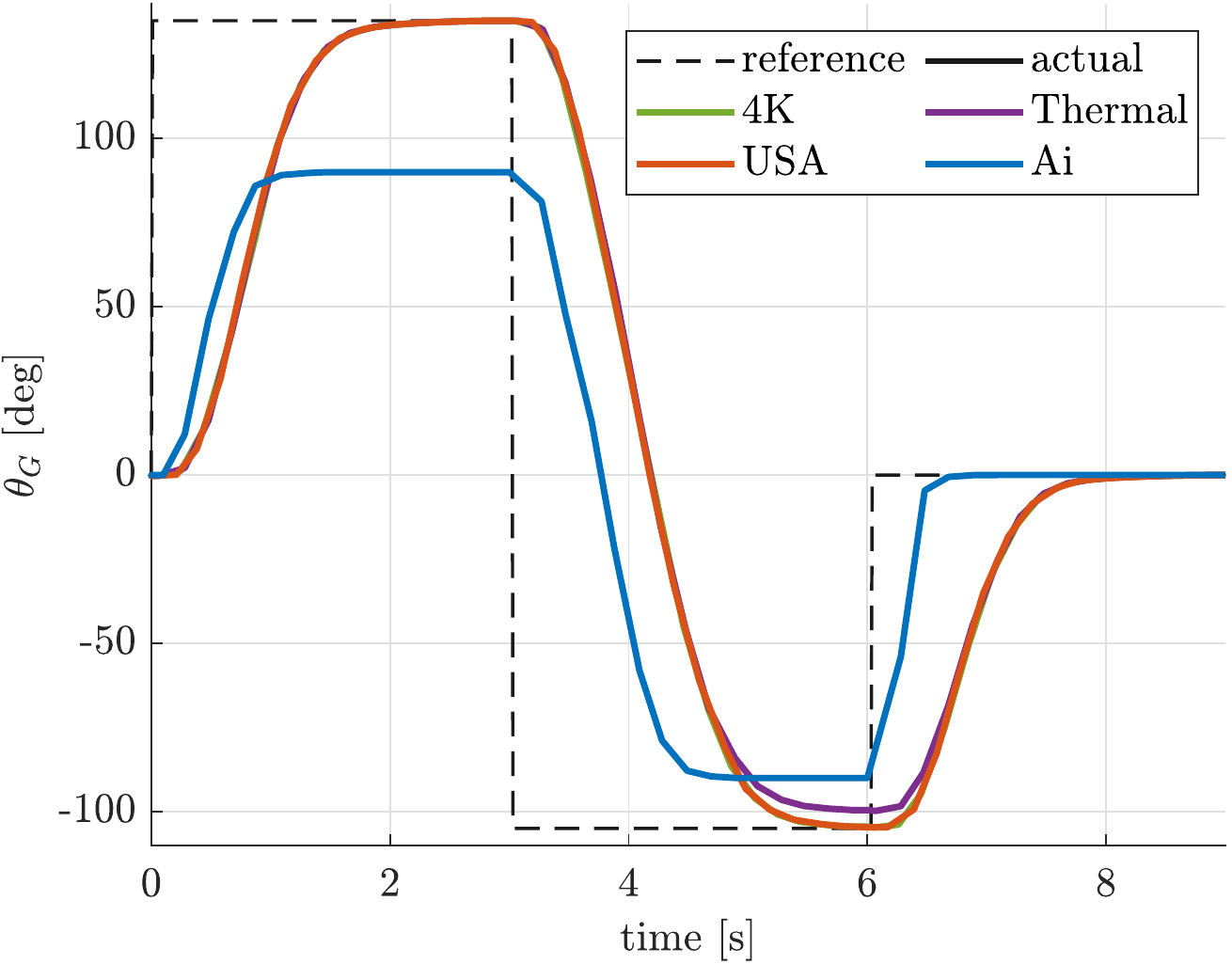}%
\label{fig:gimbal_pitch_velocity}}
\caption{Gimbal response.}
\label{fig:gimbal}
\end{figure}

All ANAFI drones have an active gimbal on which the main cameras are mounted. The gimbal can adjust its roll and pitch in orientation or angular velocity modes. Fig.~\ref{fig:gimbal} shows the responses of the gimbal in two modes for two controlled axis between the gimbal's operational limits. It is possible to observe that ANAFI Ai has the fastest response but a limited range compared to other ANAFI drones.

\begin{remark}
ANAFI 4K, Thermal and USA are equipped with similar gimbals, so their response is almost identical.
\end{remark}

\subsection{Camera}

\renewcommand{\TwoFiguresWidth}{0.9}
\begin{figure}[!b]
\centering
\subfloat[RGB image from ANAFI 4K.]{\includegraphics[width=\TwoFiguresWidth\columnwidth]{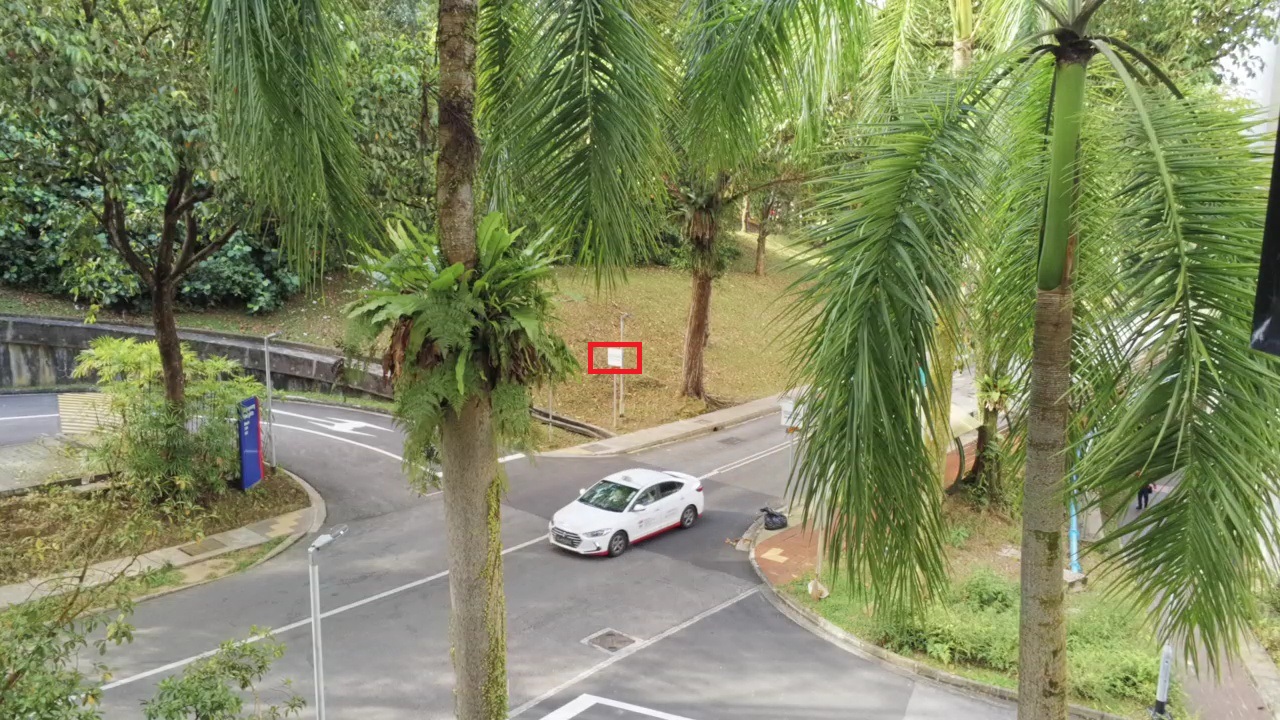}%
\label{fig:camera_4k}} \\
\subfloat[Thermal image from the ANAFI Thermal.]{\includegraphics[width=\TwoFiguresWidth\columnwidth]{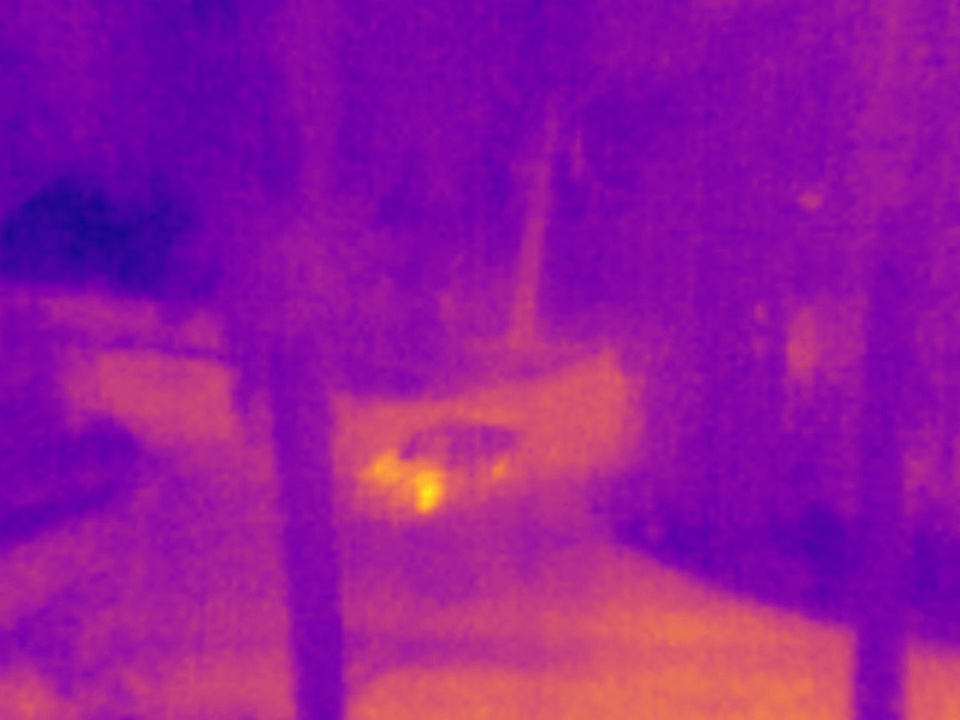}%
\label{fig:camera_thermal}} \\
\subfloat[32x zoomed RGB image from ANAFI USA.]{\includegraphics[width=\TwoFiguresWidth\columnwidth]{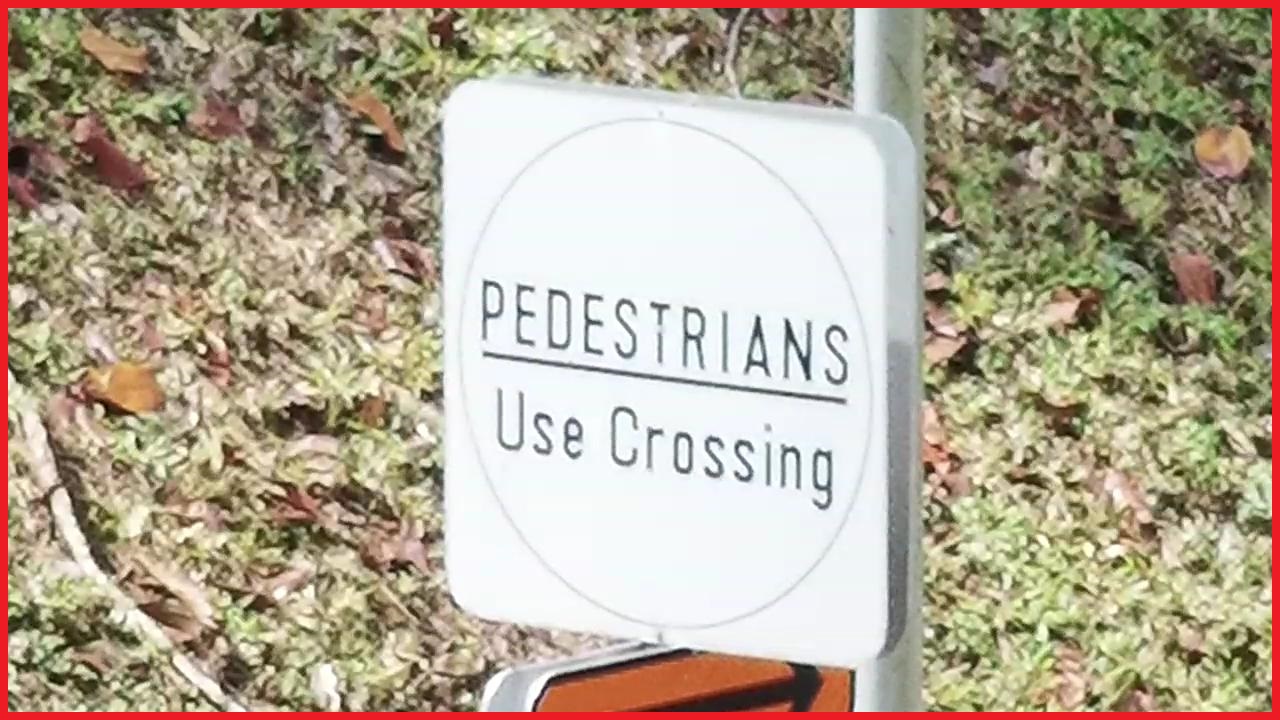}%
\label{fig:camera_usa}} \\
\subfloat[Disparity map image from ANAFI Ai.]{\includegraphics[width=\TwoFiguresWidth\columnwidth]{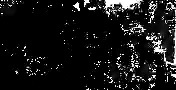}%
\label{fig:camera_ai}}
\caption{Camera features of different Anafi drones.}
\label{fig:camera}
\end{figure}

The main difference between ANAFI drones is the set of cameras they are equipped with, as summarized in Fig.~\ref{fig:camera}. The developed package allows switching online between streams from all available cameras. ANAFI~4K has one RGB front-mounted camera, which can stream live $1280 \times 720~\si{px}$ images, shown in Fig.~\ref{fig:camera_4k}. ANAFI~Thermal, besides the same RGB camera as ANAFI~4K, also has a thermal camera,  which can stream live $960 \times 720~\si{px}$ images, shown in Fig.~\ref{fig:camera_thermal}. ANAFI~USA has three front-mounted cameras: a thermal camera and two RGB wide-angle and telephoto cameras, which can stream high-details images, shown in Fig.~\ref{fig:camera_usa}, where the road sign highlighted in red in Fig.~\ref{fig:camera_4k} is zoomed in. ANAFI~Ai, besides a high-resolution RGB camera, which can stream live $1920 \times 1080~\si{px}$ images, also has a pair of frontal stereo cameras, which allow the computation of 3D environment information, like the $176 \times 90~\si{px}$ disparity map, shown in Fig.~\ref{fig:camera_ai}, where the palm leaves in the proximity are detected. Besides streaming live the video feed, all drones can shoot pictures, record videos and store them at maximum resolution on the memory card. In addition, {\tt anafi\_ros} allows downloading the stored media from the drone.

\begin{remark}
All ANAFI drones also have a down-facing grey-scale global shutter $320 \times 240~\si{px}$ camera for optical flow. However, this video stream is not accessible yet.
\end{remark}

\section{Conclusions}
\label{sec:conclusion}

This work introduces a ROS1 and ROS2 \mbox{package -- {\tt anafi\_ros} -- for} simple interfacing with the drones from the Parrot ANAFI family. The developed ROS package is hardware agnostic, allowing connecting seamlessly to four supported models. The developed package was intensively tested on the drones at maximum roll and pitch angles of $\pm 40\si{\degree}$, corresponding to the horizontal speeds above $\pm 10~\si{m/s}$, maximum vertical speed of $\pm 4~\si{m/s}$ and maximum yaw rate of $\pm 200\si{\degree/s}$. All drone models demonstrated satisfactory performance and stable response. 
We hope the developed framework will provide new opportunities for further applications of aerial robots.
\section*{Appendix}
\label{sec:appendix}

\subsection{Subscribed topics}

\noindent\textbf{\textless topic name\textgreater} (\texttt{\textless message type\textgreater}): \mbox{\textless topic description\textgreater}
\begin{itemize}
    \item \textbf{camera/command} (\hyperref[msg:CameraCommand]{\texttt{CameraCommand}}): camera zoom commands
    \item \textbf{drone/command} (\hyperref[msg:PilotingCommand]{\texttt{PilotingCommand}}): drone piloting commands
    \item \textbf{drone/moveby} (\hyperref[msg:MoveByCommand]{\texttt{MoveByCommand}}): move the drone by the given displacement and rotate by the given angle
    \item \textbf{drone/moveto} (\hyperref[msg:MoveToCommand]{\texttt{MoveToCommand}}): move the drone to the specified location
    \item \textbf{gimbal/command} (\hyperref[msg:GimbalCommand]{\texttt{GimbalCommand}}): gimbal attitude commands
\end{itemize}

\subsection{Published topics}

\noindent\textbf{\textless topic name\textgreater} (\texttt{\textless message type\textgreater}, \textless $~\si{frequency}$\textgreater)~$\in$ \{\texttt{\textless set of values\textgreater}\} / [\texttt{\textless range of values\textgreater}]: \mbox{\textless topic description\textgreater} [\si{\textless measurement~units\textgreater}]
\begin{itemize}
    \item \textbf{battery/health} (\texttt{UInt8}, $1~\si{Hz}$) $\in$ [\texttt{0}: bad, \texttt{100}: good]: battery health~[\si{\percent}]
    \item \textbf{battery/percentage} (\texttt{UInt8}, $30~\si{Hz}$) $\in$ [\texttt{0}: empty, \texttt{100}: full]: battery level~[\si{\percent}] 
    \item \textbf{battery/voltage} (\texttt{Float32}, $1~\si{Hz}$): battery voltage~[\si{V}]
    \item \textbf{camera/awb\_b\_gain} (\texttt{Float32}, $30~\si{Hz}$): camera automatic white balance (AWB) blue gain
    \item \textbf{camera/awb\_r\_gain} (\texttt{Float32}, $30~\si{Hz}$): camera automatic white balance (AWB) red gain
    \item \textbf{camera/camera\_info} (\texttt{CameraInfo}, $30~\si{Hz}$):  main camera's info
    \item \textbf{camera/exposure\_time} (\texttt{Float32}, $30~\si{Hz}$): exposure time of the main camera~[\si{s}]
    \item \textbf{camera/image} (\texttt{Image}, $30~\si{Hz}$): image from the main front camera
    \item \textbf{camera/hfov} (\texttt{Float32}, $30~\si{Hz}$): camera's horizontal field of view~[\si{\degree}]
    \item \textbf{camera/iso\_gain} (\texttt{UInt16}, $30~\si{Hz}$): camera's sensitivity gain
    \item \textbf{camera/vfov} (\texttt{Float32}, $30~\si{Hz}$): camera's vertical field of view~[\si{\degree}]
    \item \textbf{camera/zoom} (\texttt{Float32}, $5~\si{Hz}$): camera zoom level~[\si{x}]
    \item \textbf{drone/altitude} (\texttt{Float32}, $30~\si{Hz}$) $>$ \texttt{0.0}: drone's ground distance~[\si{m}]
    \item \textbf{drone/altitude\_above\_to} (\texttt{Float32}, $5~\si{Hz}$): drone's ground distance above the take-off point~[\si{m}]
    \item \textbf{drone/attitude} (\texttt{QuaternionStamped}, $30~\si{Hz}$): drone’s attitude  in north-west-up frame
    \item \textbf{drone/gps/fix} (\texttt{Bool}, $1~\si{Hz}$) $\in$ \{\texttt{true}: GPS is fixed, \texttt{false}:~GPS is not fixed\}
    \item \textbf{drone/gps/location} (\texttt{NavSatFix}, $1~\si{Hz}$): drone’s GPS location
    \item \textbf{drone/gps/satellites} (\texttt{UInt8}): number of GPS satellites
    \item \textbf{drone/rpy} (\texttt{Vector3Stamped}, $30~\si{Hz}$): drone’s roll, pitch and yaw in north-west-up frame~[\si{\degree}]
    \item \textbf{drone/speed} (\texttt{Vector3Stamped}, $30~\si{Hz}$): drone's speed in body frame~[\si{m/s}]
    \item \textbf{drone/state} (\texttt{String}, $30~\si{Hz}$) $\in$ \{`{CONNECTING}', `{LANDED}', `{TAKINGOFF}', `{HOVERING}', `{FLYING}', `{LANDING}', `{EMERGENCY}', `{DISCONNECTED}', \ldots\}: drone's state
    \item \textbf{gimbal/attitude/absolute} (\texttt{QuaternionStamped}, $5~\si{Hz}$): gimbal's attitude in north-west-up frame
    \item \textbf{home/location} (\hyperref[msg:Location]{\texttt{PointStamped}}): home location
    \item \textbf{link/quality} (\texttt{UInt8}, $30~\si{Hz}$) $\in$ [\texttt{0}: bad, \texttt{5}: good]: link quality
    \item \textbf{skycontroller/attitude} (\texttt{QuaternionStamped}, $20~\si{Hz}$): SkyController's attitude in north-west-up frame
    \item \textbf{skycontroller/command}~(\hyperref[msg:SkycontrollerCommand]{\texttt{SkycontrollerCommand}}, $100~\si{Hz}$): command from SkyController
    \item \textbf{skycontroller/rpy} (\texttt{Vector3Stamped}, $20~\si{Hz}$): SkyController's attitude in north-west-up frame~[\si{\degree}]
    \item \textbf{storage/available} (\texttt{UInt64}): available storage space~[\si{B}]
    \item \textbf{time} (\texttt{Time}, $30~\si{Hz}$): drone's local time
\end{itemize}

\subsection{Services}

\noindent\textbf{\textless service name\textgreater} (\texttt{\textless service type\textgreater}): \mbox{\textless service description\textgreater}
\begin{itemize}
    \item \textbf{camera/photo/stop} (\hyperref[srv:Photo]{\texttt{Photo}}): stop photo capture
    \item \textbf{camera/photo/take} (\hyperref[srv:Photo]{\texttt{Photo}}): take a photo
    \item \textbf{camera/recording/start} (\hyperref[srv:Recording]{\texttt{Recording}}): start video recording
    \item \textbf{camera/recording/stop} (\hyperref[srv:Recording]{\texttt{Recording}}): stop video recording
    \item \textbf{camera/reset} (\texttt{Trigger}): reset zoom level
    \item \textbf{drone/arm} (\texttt{SetBool}): \{\texttt{true}: arm the drone; \texttt{false}: disarm the drone\}
    \item \textbf{drone/calibrate} (\texttt{Trigger}): start drone's magnetometer calibration process
    \item \textbf{drone/emergency} (\texttt{Trigger}): cut out the motors
    \item \textbf{drone/halt} (\texttt{Trigger}): halt and start hovering
    \item \textbf{drone/land} (\texttt{Trigger}): take-off the drone
    \item \textbf{drone/reboot} (\texttt{Trigger}): reboot the drone
    \item \textbf{drone/rth} (\texttt{Trigger}): return home
    \item \textbf{drone/takeoff} (\texttt{Trigger}): land the drone
    \item \textbf{flightplan/pause} (\texttt{Trigger}): pause the flight plan
    \item \textbf{flightplan/start} (\hyperref[srv:FlightPlan]{\texttt{FlightPlan}}): start the flight plan based on the Mavlink file existing on the drone
    \item \textbf{flightplan/stop} (\texttt{Trigger}): stop the flight plan
    \item \textbf{flightplan/upload} (\hyperref[srv:FlightPlan]{\texttt{FlightPlan}}): upload the Mavlink file to the drone
    \item \textbf{gimbal/calibrate} (\texttt{Trigger}): start gimbal calibration
    \item \textbf{gimbal/reset} (\texttt{Trigger}): reset the reference orientation of the gimbal
    \item \textbf{home/navigate} (\texttt{SetBool}): \{\texttt{true}: start return home; \texttt{false}: stop return home\} trigger navigate home
    \item \textbf{home/set} (\hyperref[srv:Location]{\texttt{Location}}): set the custom home location
    \item \textbf{skycontroller/offboard} (\texttt{SetBool}): \{\texttt{true}: switch to offboard control; \texttt{false}: switch to manual control\} change control mode
    \item \textbf{storage/download} (\texttt{SetBool}): \{\texttt{true}: delete media after download; \texttt{false}: otherwise\} download media from the drone
    \item \textbf{storage/format} (\texttt{Trigger}): format removable storage
\end{itemize}

\subsection{Parameters}

\noindent\textbf{\textless parameter name\textgreater} (\texttt{\textless parameter type\textgreater})~:= \texttt{\textless default value\textgreater} $\in$ \{\texttt{\textless set of values\textgreater}\} / [\texttt{\textless range of values\textgreater}]: \mbox{\textless parameter description\textgreater}~[\si{\textless measurement~units\textgreater}]
\begin{itemize}
    \item \textbf{camera/autorecord} (\texttt{bool}) := \texttt{false} $\in$ \{\texttt{true}: enabled; \texttt{false}: disabled\}: auto record at take-off
    \item \textbf{camera/ev\_compensation} (\texttt{int}) := \texttt{9} $\in$ \{\texttt{0}:~$-3.00$; \texttt{3}:~$-2.00$; \texttt{6}:~$-1.00$; \texttt{9}:~$0.00$; \texttt{12}:~$1.00$; \texttt{15}:~$2.00$; \texttt{18}:~$3.00$\}: camera exposure compensation~[\si{EV}]
    \item \textbf{camera/hdr} (\texttt{bool}) := \texttt{true} $\in$ \{\texttt{true}: enabled; \texttt{false}:~disabled\}: high dynamic range~(HDR) mode
    \item \textbf{camera/max\_zoom\_speed} (\texttt{float}) := \texttt{10.0} $\in$ [\texttt{0.1}, \texttt{10.0}]: maximum zoom speed~[$\tan(\si{\degree})$~\si{/s}]
    \item \textbf{camera/mode} (\texttt{int}) := \texttt{0} $\in$ \{\texttt{0}: camera in recording mode; \texttt{1}:~camera~in~photo~mode\}: camera mode
    \item \textbf{camera/relative} (\texttt{bool}) := \texttt{false} $\in$ \{\texttt{true}: commands relative to the camera pitch; \texttt{false}: otherwise\} 
    \item \textbf{camera/rendering} (\texttt{int}) := 0 $\in$ \{\texttt{0}:~visible; \texttt{1}:~thermal; \texttt{2}:~blended\}: thermal image rendering mode (\texttt{1} and \texttt{2} supported only by ANAFI Thermal and ANAFI USA)
    \item \textbf{camera/streaming} (\texttt{int}) := \texttt{0} $\in$ \{\texttt{0}: minimize latency with average reliability (best for piloting); \texttt{1}: maximize reliability with an average latency; \texttt{2}: maximize reliability using a frame-rate decimation\}: streaming mode
    \item \textbf{camera/style} (\texttt{int}) := \texttt{0} $\in$ \{\texttt{0}: natural look; \texttt{1}:~flat and desaturated images, best for post-processing; \texttt{2}:~intense~--~bright colors, warm shade, high contrast; \texttt{3}:~pastel~--~soft colors, cold shade, low contrast\}: images style
    \item \textbf{drone/banked\_turn} (\texttt{bool}) := \texttt{true} $\in$ \{\texttt{true}: enabled; \texttt{false}: disabled\}: banked turn
    \item \textbf{drone/max\_altitude} (\texttt{float}) := \texttt{2.0} $\in$ [\texttt{0.5}, \texttt{4000.0}]: maximum altitude~[\si{m}]
    \item \textbf{drone/max\_distance} (\texttt{float}) := \texttt{10.0} $\in$ [\texttt{10.0}, \texttt{4000.0}]: maximum distance~[\si{m}]
    \item \textbf{drone/max\_horizontal\_speed} (\texttt{float}) := \texttt{1.0} $\in$ [\texttt{0.1},~\texttt{15.0}]: maximum horizontal speed~[\si{m/s}]
    \item \textbf{drone/max\_pitch\_roll} (\texttt{float}) := \texttt{10.0} $\in$ [\texttt{1.0}, \texttt{40.0}]: maximum pitch and roll angle~[\si{\degree}]
    \item \textbf{drone/max\_pitch\_roll\_rate} (\texttt{float}) := \texttt{200.0} $\in$ [\texttt{40.0},~\texttt{300.0}]: maximum pitch and roll rotation speed~[\si{\degree/s}]
    \item \textbf{drone/max\_vertical\_speed} (\texttt{float}) := \texttt{1.0} $\in$ [\texttt{0.1}, \texttt{4.0}]: maximum vertical speed~[\si{m/s}]
    \item \textbf{drone/max\_yaw\_rate} (\texttt{float}) := \texttt{180.0} $\in$ [\texttt{3.0},~\texttt{200.0}]: maximum yaw rotation speed~[\si{\degree/s}]
    \item \textbf{drone/model} (\texttt{string}) := $\in$ \{`4k', `thermal', `usa', `ai', `unknown'\}: drone's model
    \item \textbf{gimbal/max\_speed} (\texttt{float}) := \texttt{180.0} $\in$ [\texttt{1.0},~\texttt{180.0}]: maximum gimbal speed~[\si{\degree/s}]
    \item \textbf{home/autotrigger} (\texttt{bool}) := \texttt{true} $\in$ \{\texttt{true}: enabled; \texttt{false}:~disabled\}: auto trigger return-to-home
    \item \textbf{home/ending\_behavior} (\texttt{int}) := \texttt{1} $\in$ \{\texttt{0}: land; \texttt{1}: hover\}: return-to-home ending behavior
    \item \textbf{home/min\_altitude} (\texttt{float}) := \texttt{20.0} $\in$ [\texttt{20.0}, \texttt{100.0}]: return-to-home minimum altitude [\si{m}]
    \item \textbf{home/precise} (\texttt{bool}) := \texttt{true} $\in$ \{\texttt{true}: enabled; \texttt{false}:~disabled\}: precise return-to-home
    \item \textbf{home/type} (\texttt{int}) := \texttt{4} $\in$ \{\texttt{1}: take-off location; \texttt{3}:~user-set custom location; \texttt{4}:~pilot location\}: home type for return-to-home
    \item \textbf{storage/download\_folder} (\texttt{string}) := ``$~\sim$/Pictures/Anafi'': path to the download folder
\end{itemize}

\subsection{Custom messages}

\begin{enumerate}[label=\textbf{\textbullet}]
    \item \textbf{\texttt{CameraCommand}} \label{msg:CameraCommand}
    \begin{itemize}[label=\textbf{--}]
        \item \texttt{Header} \textbf{header}: header of the message
        \item \texttt{uint8} \textbf{mode} $\in$ \{\texttt{0}: level; \texttt{1}: velocity\}: control mode
        \item \texttt{float32} \textbf{zoom}: zoom command [\si{x}] / [\si{x/s}]
    \end{itemize}
    \item \textbf{\texttt{GimbalCommand}} \label{msg:GimbalCommand}
    \begin{itemize}[label=\textbf{--}]
        \item \texttt{Header} \textbf{header}: header of the message
        \item \texttt{uint8} \textbf{mode} $\in$ \{\texttt{0}: position; \texttt{1}: velocity\}: control mode
        \item \texttt{uint8} \textbf{frame} $\in$ \{\texttt{0}:~none; \texttt{1}:~relative; \texttt{2}:~absolute\}: gimbal's frame of reference
        \item \texttt{float32} \textbf{roll}: roll command [\si{\degree}] / [\si{\degree/s}]
        \item \texttt{float32} \textbf{pitch}: pitch command [\si{\degree}] / [\si{\degree/s}]
        \item \texttt{float32} \textbf{yaw}: pitch command [\si{\degree}] / [\si{\degree/s}]
    \end{itemize}
    \item \textbf{\texttt{MoveByCommand}} \label{msg:MoveByCommand}
    \begin{itemize}[label=\textbf{--}]
        \item \texttt{Header} \textbf{header}: header of the message
        \item \texttt{float32} \textbf{dx}: $x$ displacement [\si{m}]
        \item \texttt{float32} \textbf{dy}: $y$ displacement [\si{m}]
        \item \texttt{float32} \textbf{dz}: $z$ displacement [\si{m}]
        \item \texttt{float32} \textbf{dyaw}: yaw displacement [\si{\degree}]
    \end{itemize}
    \item \textbf{\texttt{MoveToCommand}} \label{msg:MoveToCommand}
    \begin{itemize}[label=\textbf{--}]
        \item \texttt{Header} \textbf{header}: header of the message
        \item \texttt{float64} \textbf{latitude}: latitude [\si{\degree}]
        \item \texttt{float64} \textbf{longitude}: longitude [\si{\degree}]
        \item \texttt{float64} \textbf{altitude}: altitude [\si{m}]
        \item \texttt{float32} \textbf{heading}: heading w.r.t. North [\si{\degree}]
        \item \texttt{uint8} \textbf{orientation\_mode} $\in$ \{\texttt{0}:~none; \texttt{1}:~to target; \texttt{2}:~heading start; \texttt{3}:~heading during\}: orientation mode
    \end{itemize}
    \item \textbf{\texttt{PilotingCommand}} \label{msg:PilotingCommand}
    \begin{itemize}[label=\textbf{--}]
        \item \texttt{Header} \textbf{header}: header of the message
        \item \texttt{float32} \textbf{roll}: roll angle [\si{\degree}]
        \item \texttt{float32} \textbf{pitch}: pitch angle [\si{\degree}]
        \item \texttt{float32} \textbf{yaw}: yaw rate [\si{\degree/s}]
        \item \texttt{float32} \textbf{gaz}: vertical velocity [\si{m/s}]
    \end{itemize}
    \item \textbf{\texttt{SkycontrollerCommand}} \label{msg:SkycontrollerCommand}
    \begin{itemize}[label=\textbf{--}]
        \item \texttt{Header} \textbf{header}: header of the message
        \item \texttt{int8} \textbf{x} $\in$ [\texttt{-100}, \texttt{100}]: $x$-axis~[\si{\percent}]
        \item \texttt{int8} \textbf{y} $\in$ [\texttt{-100}, \texttt{100}]: $y$-axis~[\si{\percent}]
        \item \texttt{int8} \textbf{z} $\in$ [\texttt{-100}, \texttt{100}]: $z$-axis~[\si{\percent}]
        \item \texttt{int8} \textbf{yaw} $\in$ [\texttt{-100}, \texttt{100}]: yaw-axis~[\si{\percent}]
        \item \texttt{int8} \textbf{camera} $\in$ [\texttt{-100}, \texttt{100}]: camera-axis~[\si{\percent}]
        \item \texttt{int8} \textbf{zoom} $\in$ [\texttt{-100}, \texttt{100}]: zoom-axis~[\si{\percent}]
        \item \texttt{bool} \textbf{return\_home} $\in$ \{\texttt{true}: pressed; \texttt{false}: not pressed\}: return-to-home (front top) button
        \item \texttt{bool} \textbf{takeoff\_land} $\in$ \{\texttt{true}: pressed; \texttt{false}: not pressed\}: take-off/land (front bottom) button
        \item \texttt{bool} \textbf{reset\_camera} $\in$ \{\texttt{true}: pressed; \texttt{false}: not pressed\}: reset camera (back left) button
        \item \texttt{bool} \textbf{reset\_zoom} $\in$ \{\texttt{true}: pressed; \texttt{false}: not pressed\}: reset zoom (back right) button
    \end{itemize}
\end{enumerate}

\subsection{Custom services}

\begin{itemize}
    \item \textbf{\texttt{FlightPlan}} \label{srv:FlightPlan}
    \begin{itemize}
        \item \texttt{string} \textbf{file}: path to the flight plan file on local computer
        \item \texttt{string} \textbf{uid}: flight plan UID in drone's directory
    \end{itemize}
    \item \textbf{\texttt{Location}} \label{srv:Location}
    \begin{itemize}
        \item \texttt{float64} \textbf{latitude}: latitude [\si{\degree}]
        \item \texttt{float64} \textbf{longitude}: longitude [\si{\degree}]
        \item \texttt{float64} \textbf{altitude}: altitude [\si{m}]
    \end{itemize}
    \item \textbf{\texttt{PilotedPOI}} \label{srv:PilotedPOI}
    \begin{itemize}
        \item \texttt{float64} \textbf{latitude}: latitude to look at [\si{\degree}]
        \item \texttt{float64} \textbf{longitude}: longitude to look at [\si{\degree}]
        \item \texttt{float64} \textbf{altitude}: altitude to look at [\si{m}]
        \item \texttt{bool} \textbf{locked\_gimbal} $\in$ \{\texttt{true}: gimbal is locked on the point of interest, \texttt{false}: gimbal is freely controllable\}: gimbal is locked
    \end{itemize}
    \item \textbf{\texttt{Photo}} \label{srv:Photo} $\to$ \texttt{string} \textbf{media\_id}: media id
    \begin{itemize}
        \item \texttt{uint8} \textbf{mode} $\in$ \{\texttt{0}: single shot; \texttt{1}:~bracketing -- burst of frames with a different exposure; \texttt{2}:~burst of frames; \texttt{3}:~time-lapse -- frames at a regular time interval; \texttt{4}:~GPS-lapse -- frames at a regular GPS position interval\}: photo mode 
        \item \texttt{uint8} \textbf{photo\_format} $\in$ \{\texttt{0}: full resolution, not dewarped; \texttt{1}: rectilinear projection, dewarped\}: photo format
        \item \texttt{uint8} \textbf{file\_format} $\in$ \{\texttt{0}: jpeg; \texttt{1}: dng; \texttt{2}:~jpeg and dng\}: file format
    \end{itemize}
    \item \textbf{\texttt{Recording}} \label{srv:Recording} $\to$ \texttt{string} \textbf{media\_id}: media id
    \begin{itemize}
        \item \texttt{uint8} \textbf{mode} $\in$ \{\texttt{0}: standard; \texttt{1}:~hyperlapse; \texttt{2}:~slow motion; \texttt{3}:~high-framerate\}: video recording mode
    \end{itemize}
\end{itemize}

\section*{Acknowledgment}

The author thanks the support team of Parrot for their assistance.

\bibliographystyle{IEEEtran}
\bibliography{References}

\begin{thebibliography}{10}
\providecommand{\url}[1]{#1}
\csname url@samestyle\endcsname
\providecommand{\newblock}{\relax}
\providecommand{\bibinfo}[2]{#2}
\providecommand{\BIBentrySTDinterwordspacing}{\spaceskip=0pt\relax}
\providecommand{\BIBentryALTinterwordstretchfactor}{4}
\providecommand{\BIBentryALTinterwordspacing}{\spaceskip=\fontdimen2\font plus
\BIBentryALTinterwordstretchfactor\fontdimen3\font minus
  \fontdimen4\font\relax}
\providecommand{\BIBforeignlanguage}[2]{{%
\expandafter\ifx\csname l@#1\endcsname\relax
\typeout{** WARNING: IEEEtran.bst: No hyphenation pattern has been}%
\typeout{** loaded for the language `#1'. Using the pattern for}%
\typeout{** the default language instead.}%
\else
\language=\csname l@#1\endcsname
\fi
#2}}
\providecommand{\BIBdecl}{\relax}
\BIBdecl

\bibitem{Tranzatto2022SR}
M.~Tranzatto, T.~Miki, M.~Dharmadhikari, L.~Bernreiter, M.~Kulkarni,
  F.~Mascarich, O.~Andersson, S.~Khattak, M.~Hutter, R.~Siegwart, and
  K.~Alexis, ``{CERBERUS in the DARPA Subterranean Challenge},'' \emph{Science
  Robotics}, vol.~7, no.~66, p. eabp9742, 2022.

\bibitem{Suarez2022ICUAS}
A.~Suarez, R.~Salmoral, A.~Garofano-Soldado, G.~Heredia, and A.~Ollero,
  ``{Aerial Device Delivery for Power Line Inspection and Maintenance},'' in
  \emph{2022 International Conference on Unmanned Aircraft Systems (ICUAS)},
  2022, pp. 30--38.

\bibitem{Zhang2022Nature}
K.~Zhang, P.~Chermprayong, F.~Xiao, D.~Tzoumanikas, B.~Dams, S.~Kay, B.~Kocer,
  A.~Burns, L.~Orr, C.~Choi, D.~Darekar, W.~Li, S.~Hirschmann, V.~Soana,
  S.~Ngah, S.~Sareh, A.~Choubey, L.~Margheri, V.~Pawar, and M.~Kovac, ``{Aerial
  additive manufacturing with multiple autonomous robots},'' \emph{Nature},
  vol. 609, pp. 709--717, 09 2022.

\bibitem{Valero2021EO}
M.~M. Valero, S.~Verstockt, B.~Butler, D.~Jimenez, O.~Rios, C.~Mata, L.~Queen,
  E.~Pastor, and E.~Planas, ``{Thermal Infrared Video Stabilization for Aerial
  Monitoring of Active Wildfires},'' \emph{IEEE Journal of Selected Topics in
  Applied Earth Observations and Remote Sensing}, vol.~14, pp. 2817--2832,
  2021.

\bibitem{Chen2022EO}
L.~Chen, Z.~Fang, and Y.~Fu, ``{Consistency-Aware Map Generation at Multiple
  Zoom Levels Using Aerial Image},'' \emph{IEEE Journal of Selected Topics in
  Applied Earth Observations and Remote Sensing}, vol.~15, pp. 5953--5966,
  2022.

\bibitem{Pham2022RAL}
H.~X. Pham, A.~Sarabakha, M.~Odnoshyvkin, and E.~Kayacan, ``{PencilNet:
  Zero-Shot Sim-to-Real Transfer Learning for Robust Gate Perception in
  Autonomous Drone Racing},'' \emph{IEEE Robotics and Automation Letters},
  vol.~7, no.~4, pp. 11\,847--11\,854, 2022.

\bibitem{Foehn2022SR}
P.~Foehn, E.~Kaufmann, A.~Romero, R.~Penicka, S.~Sun, L.~Bauersfeld,
  T.~Laengle, G.~Cioffi, Y.~Song, A.~Loquercio, and D.~Scaramuzza,
  ``{Agilicious: Open-source and open-hardware agile quadrotor for vision-based
  flight},'' \emph{Science Robotics}, vol.~7, no.~67, p. eabl6259, 2022.

\bibitem{dji}
\BIBentryALTinterwordspacing
``{DJI Developer},'' 2023. [Online]. Available: \url{https://developer.dji.com}
\BIBentrySTDinterwordspacing

\bibitem{ryze}
\BIBentryALTinterwordspacing
``{Tello-Python},'' 2019. [Online]. Available:
  \url{https://github.com/dji-sdk/Tello-Python}
\BIBentrySTDinterwordspacing

\bibitem{parrot}
\BIBentryALTinterwordspacing
``{Parrot Drone SDK},'' 2023. [Online]. Available:
  \url{https://developer.parrot.com/docs/index.html}
\BIBentrySTDinterwordspacing

\bibitem{bitcraze}
\BIBentryALTinterwordspacing
``{SDL Bitcraze},'' 2023. [Online]. Available:
  \url{https://www.bitcraze.io/documentation/repository/aideck-gap8-examples/master/getting-started/sdk/}
\BIBentrySTDinterwordspacing

\bibitem{bebop-autonomy}
\BIBentryALTinterwordspacing
``bebop\_autonomy,'' 2018. [Online]. Available:
  \url{https://github.com/AutonomyLab/bebop\_autonomy}
\BIBentrySTDinterwordspacing

\bibitem{tello-ros}
\BIBentryALTinterwordspacing
``tello\_ros,'' 2022. [Online]. Available:
  \url{https://github.com/clydemcqueen/tello_ros}
\BIBentrySTDinterwordspacing

\bibitem{Preiss2017ICRA}
J.~A. Preiss, W.~Honig, G.~S. Sukhatme, and N.~Ayanian, ``{Crazyswarm: A large
  nano-quadcopter swarm},'' in \emph{2017 IEEE International Conference on
  Robotics and Automation (ICRA)}, 2017, pp. 3299--3304.

\bibitem{Mur2015TRo}
R.~Mur-Artal, J.~M.~M. Montiel, and J.~D. Tardós, ``{ORB-SLAM: A Versatile and
  Accurate Monocular SLAM System},'' \emph{IEEE Transactions on Robotics},
  vol.~31, no.~5, pp. 1147--1163, 2015.

\bibitem{Tallamraju2019RAL}
R.~Tallamraju, E.~Price, R.~Ludwig, K.~Karlapalem, H.~H. Bülthoff, M.~J.
  Black, and A.~Ahmad, ``{Active Perception Based Formation Control for
  Multiple Aerial Vehicles},'' \emph{IEEE Robotics and Automation Letters},
  vol.~4, no.~4, pp. 4491--4498, 2019.

\bibitem{Zhou2021TRo}
B.~Zhou, J.~Pan, F.~Gao, and S.~Shen, ``{RAPTOR: Robust and Perception-Aware
  Trajectory Replanning for Quadrotor Fast Flight},'' \emph{IEEE Transactions
  on Robotics}, vol.~37, no.~6, pp. 1992--2009, 2021.

\bibitem{Torrente2021RAL}
G.~Torrente, E.~Kaufmann, P.~Föhn, and D.~Scaramuzza, ``{Data-Driven MPC for
  Quadrotors},'' \emph{IEEE Robotics and Automation Letters}, vol.~6, no.~2,
  pp. 3769--3776, 2021.

\bibitem{Sarabakha2020TFS}
A.~Sarabakha and E.~Kayacan, ``{Online Deep Fuzzy Learning for Control of
  Nonlinear Systems Using Expert Knowledge},'' \emph{IEEE Transactions on Fuzzy
  Systems}, vol.~28, no.~7, pp. 1492--1503, 2020.

\bibitem{Sarabakha2016CDC}
A.~Sarabakha and E.~Kayacan, ``{Y6 Tricopter Autonomous Evacuation in an Indoor
  Environment Using Q-Learning Algorithm},'' in \emph{2016 IEEE 55th Conference
  on Decision and Control (CDC)}, Dec 2016, pp. 5992--5997.

\bibitem{Mistler2001ROMAN}
V.~Mistler, A.~Benallegue, and N.~M'Sirdi, ``Exact linearization and
  noninteracting control of a 4 rotors helicopter via dynamic feedback,'' in
  \emph{Proceedings 10th IEEE International Workshop on Robot and Human
  Interactive Communication (ROMAN)}, 2001, pp. 586--593.

\end{thebibliography}

\end{document}